\documentclass[letterpaper]{article} 
\usepackage{aaai2026}  
\usepackage{times}  
\usepackage{helvet}  
\usepackage{courier}  
\usepackage[hyphens]{url}  
\usepackage{graphicx} 
\usepackage{natbib}  
\usepackage{caption} 
\usepackage{algorithm}
\usepackage{algorithmic}
\usepackage{fontawesome}
\usepackage{amsmath}
\usepackage{newfloat}
\usepackage{listings}
\DeclareCaptionStyle{ruled}{labelfont=normalfont,labelsep=colon,strut=off} 
\floatstyle{ruled}
\newfloat{listing}{tb}{lst}{}
\floatname{listing}{Listing}
\usepackage{booktabs} 
\usepackage[table]{xcolor}
\usepackage{multirow}
\usepackage{pifont}
\usepackage{array}
\newcolumntype{C}[1]{>{\centering\arraybackslash}p{#1}}
\newcolumntype{L}[1]{>{\raggedright\arraybackslash}p{#1}}

\renewcommand{\footnotemark}{\relax} 

\title{LiViBench: An Omnimodal Benchmark for Interactive Livestream Video Understanding}
\author{
    Xiaodong Wang\textsuperscript{\rm 1},
    Langling Huang\textsuperscript{\rm 1},
    Zhirong Wu\textsuperscript{\rm 1},
    Xu Zhao\textsuperscript{\rm 2},
    Teng Xu\textsuperscript{\rm 2*}
    Xuhong Xia\textsuperscript{\rm 2},
    Peixi Peng\textsuperscript{\rm 1*}\footnote{*Corresponding author.}
}
\affiliations{
    \textsuperscript{\rm 1}School of Electronic and Computer Engineering, Peking University\\

    \textsuperscript{\rm 2}Douyin Group\\
    \{wangxiaodong21s@stu., pxpeng@\}pku.edu.cn
}

\usepackage{bibentry}

\usepackage{etoolbox}
\pretocmd{\appendix}{\onecolumn}{}{}

\begin{document}

\maketitle

\begin{abstract}
The development of multimodal large language models (MLLMs) has advanced general video understanding. However, existing video evaluation benchmarks primarily focus on non-interactive videos, such as movies and recordings. To fill this gap, this paper proposes the first omnimodal benchmark for interactive livestream videos, LiViBench. It features a diverse set of 24 tasks, highlighting the perceptual, reasoning, and livestream-specific challenges. To efficiently construct the dataset, we design a standardized semi-automatic annotation workflow that incorporates the human-in-the-loop at multiple stages. The workflow leverages multiple MLLMs to form a multi-agent system for comprehensive video description and uses a seed-question-driven method to construct high-quality annotations. All interactive videos in the benchmark include audio, speech, and real-time comments modalities. To enhance models' understanding of interactive videos, we design tailored two-stage instruction-tuning and propose a Video-to-Comment Retrieval (VCR) module to improve the model's ability to utilize real-time comments. Based on these advancements, we develop LiVi-LLM-7B, an MLLM with enhanced knowledge of interactive livestreams. Experiments show that our model outperforms larger open-source models with up to 72B parameters, narrows the gap with leading proprietary models on LiViBench, and achieves enhanced performance on general video benchmarks, including VideoMME, LongVideoBench, MLVU, and VideoEval-Pro.
\end{abstract}

\begin{links}
    \link{Code}{https://github.com/Wang-Xiaodong1899/LiViBench}
\end{links}

\section{Introduction}
The rapid advancement of Multimodal Large Language Models (MLLMs) has driven significant progress in general video understanding~\cite{comanici2025gemini25,guo2025seed15vl}. Challenging benchmarks can drive rapid progress in model capabilities. Accordingly, many recent efforts have advanced the video understanding benchmark ecosystem by introducing more challenging tasks and longer videos~\cite{fu2025videomme,wang2024lvbench}. However, existing video benchmarks primarily focus on non-interactive content, such as movies, recordings, and short videos, and lack coverage of interactive videos, such as livestreams that involve rich and frequent interactions between streamers and audiences. Given the growing prevalence of livestreams in online video consumption, enhancing models’ ability to understand such interactive content has become increasingly important.

Unlike general videos, livestreams are inherently interactive, emphasizing real-time engagement between streamers and their audiences. This interaction includes, but is not limited to, gift-giving, live conversations, audiences' real-time comments, and multi-person co-streaming. These interactive features demonstrate the unique characteristics of livestreams. Despite recent advancements in MLLMs, it remains unclear how effectively these models can comprehend these types of interactive videos.

To address the lack of interactive video content in existing video understanding benchmarks and evaluate the comprehension capabilities of MLLMs on livestream videos, we introduce LiViBench, the first omnimodal interactive video benchmark. The benchmark focuses on human-centered themes and covers 9 vertical domains of interactive livestream (e.g., chatting and singing), and the task taxonomy comprehensively includes 24 distinct tasks (e.g., multi-person interaction and behavior reasoning). These tasks span a broad range of categories, including general perception and reasoning, knowledge-based question answering, and livestream-specific tasks that highlight the interactive characteristics of livestream scenes. LiViBench comprises 3,168 livestream videos with durations ranging from 14 seconds to 33 minutes, along with 3,175 high-quality multiple-choice questions. At the same time, we introduce rich and heterogeneous data to enable a more comprehensive evaluation, covering audio, speech, and comments.

Constructing a video benchmark poses challenges to video annotation and question quality. While prior work has either lacked transparency~\cite{fu2025videomme,wu2024longvideobench} or relied entirely on automated annotations~\cite{han2025videoespresso}, we design a standardized semi-automatic annotation workflow, and incorporate human-in-the-loop at multiple stages. To reduce the bias introduced by MLLMs during automatic annotation, we first use multiple MLLMs to build a multi-agent system that comprehensively describes the video. For each task, we construct a seed question library through automatic generation using proprietary models, followed by human revision and augmentation. Using a seed-question–driven strategy, models generate candidate questions for each video, which are then screened and refined by humans. Both models and humans provide answers to the selected questions. Finally, human annotators conduct thorough quality control on the resulting multi-choice QA set to ensure clarity, correctness, and relevance.

Based on the constructed LiViBench, we perform a comprehensive evaluation of both state-of-the-art proprietary models and open-source models. Preliminary experiments reveal that proprietary models (e.g., GPT-4o and Gemini-2.5-Pro) exhibit notable limitations in understanding interactive videos, and large-scale open-source models also demonstrate constrained performance. These limitations are likely due to the lack of instruction-tuning datasets specifically designed for interactive videos. To address this, we construct an instruction-tuning dataset comprising 37,953 machine-annotated and 11,180 manually annotated samples for interactive videos. We further propose a tailored two-stage instruction tuning strategy to fully leverage the training data.

In the livestream video domain, real-time comment is a unique and important modality. The sheer volume of these comments poses significant challenges for both the input context length and the information extraction abilities of MLLMs. To evaluate their impact on video understanding, we incorporate the real-time comments into each task. To better leverage these comments, we propose a Video-to-Comment Retrieval (VCR) module that retrieves relevant comments using video features. Together with the tailored instruction tuning, we develop LiVi-LLM-7B, a video understanding model enriched with enhanced knowledge of interactive livestreams. Experimental results demonstrate that our model outperforms larger open-source models with up to 72B parameters and narrows the gap to the best proprietary models on LiViBench. It also shows strong generalization across general video benchmarks of varying lengths, including Video-MME, LongVideoBench, MLVU, and VideoEval-Pro. The contributions are as follows:
\begin{itemize}

    \item We propose the first omnimodal benchmark specifically designed for interactive livestream videos, LiViBench, with audio, speech, and comment modalities. The comprehensive evaluation shows that some proprietary models (e.g., GPT-4o and Gemini) have limited performance on this new benchmark.

    \item We propose a standardized semi-automatic annotation workflow that introduces human-in-the-loop in multiple stages. The multi-agent mechanism and seed question library are introduced to efficiently construct high-quality evaluation data and instruction-tuning data.

    \item We design tailored instruction tuning and a video-to-comment module to build the comprehension-enhanced model: LiVi-LLM-7B. It outperforms larger open-source models, including those with up to 72B parameters, and narrows the gap with the best proprietary models.
\end{itemize}

\section{Related Work}

\subsection{Multi-Modal Large Language Models}
Multimodal Large Language Models (MLLMs)~\cite{hurst2024gpt4o,anthropic3Claude,comanici2025gemini25} have achieved significant advancements in video understanding tasks. These models typically treat video as a sequence of images and connect the output of the visual encoder to a large language model (LLM) through a modality alignment module, enabling visual content understanding and reasoning. The architecture and training paradigms of video understanding systems continue to evolve. Early works incorporate Q-Former~\cite{li2023blip} to extract informative video features while reducing the number of visual tokens~\cite{zhang2023videollama,ren2024timechat,wang2024internvideo2}. Then, some works adopt a simpler approach by using an MLP to directly project the processed visual features into the feature space of the LLM~\cite{zhang2024llavanextvideo,liu2025nvila,maaz2023video-chatgpt}. Recently, Qwen2.5-VL~\cite{bai2025qwen25vl} fuses adjacent frames at the input stage and further compresses encoded multiple visual tokens into a single token, which is then connected to the language model via MLP. To enhance temporal awareness and achieve event-level localization, TimeChat~\cite{ren2024timechat} introduces explicit temporal textual prompts, TimeSuite~\cite{zeng2025timesuite} incorporates the TAPE module to capture temporal structures, and Qwen2.5-VL utilizes the MRoPE to model inter-frame temporal relationships. Regarding training data, many works adopt a hybrid data training strategy. For instance, InternVL2.5~\cite{chen2024internvl25} and Qwen2.5-VL are trained on a combination of single images, multi-frame image sequences, and videos. Additionally, post-training techniques are widely used to improve reasoning performance~\cite{zhu2025internvl3,wang2025leanpo,wang2025open}.

Moreover, researchers are actively exploring omnimodal models capable of processing text, images, videos, and audio, e.g.,~\cite{cheng2024videollama2},~\cite{xu2025qwen25omni},~\cite{yao2024minicpm}, and~\cite{liu2025ola}, aiming to further expand the perceptual capabilities of omnimodal systems. Despite the strong performance of the aforementioned models in general video understanding tasks, their adaptability to live streaming scenarios remains underexplored. Related model Kwai-Keye~\cite{team2025keye} is primarily optimized for non-interactive short videos and struggles to process interactive livestream videos. Therefore, our work targets interactive livestream video understanding. We introduce an omnimodal benchmark for interactive videos and develop a model enriched with interactive knowledge.

\subsection{Multi-Modal Video Benchmarks}
The introduction of various benchmarks has promoted the development of MLLMs. Existing video benchmarks mainly focus on general video understanding tasks, such as short video understanding tasks~\cite{msrvtt,yu2019activitynet,li2024mvbench,fang2024mmbench,hong2025motionbench,liu2024tempcompass,shangguan2024tomato}, video temporal grounding tasks~\cite{gao2017charades,lei2021qvhighlight,rohrbach2014tacos}, video reasoning tasks ~\cite{hu2025videommmu,zhao2025mmvu,rasheed2025videomathqa,zhu2025mmr,han2025videoespresso}, long video understanding tasks~\cite{fu2025videomme,wang2024lvbench,zhou2024mlvu,ma2025videoeval}, in order to perceive the real world more comprehensively, some audio-visual benchmarks have been proposed~\cite{li2022musicavqa,geng2025longvale,hong2025worldsense}.

Some other benchmarks focus on specific domains, such as first-person videos~\cite{mangalam2023egoschema}, cinematic language~\cite{liu2025shotbench}, video comments~\cite{lei2025godbench}, video quality and aesthetics~\cite{jia2025omnivqa}, content moderation on short video platforms~\cite{lu2025vlmaspolicy}, and offline short videos from Kuaishou~\cite{team2025kwai}.
Despite the emergence of benchmarks, current video understanding benchmarks mainly focus on non-interactive videos. With the rapid growth of social media platforms such as Instagram Live and TikTok Live, livestream videos have become increasingly prevalent, bringing with them complex multimodal tasks that challenge existing model capabilities. A benchmark that can evaluate the ability of MLLMs to understand such data is needed. Therefore, we propose LiViBench, the first omnimodal livestream video benchmark.

\section{Method}

\subsection{Video Curation}
To evaluate the capability of MLLMs in understanding interactive livestream videos, we curate a diverse and comprehensive dataset from publicly accessible livestream videos. This dataset comprises synchronized multi-modal data, including video, audio, speech, and user comments, which provides rich contextual information. To facilitate human-centered video comprehension, reasoning, and knowledge-based question answering, we focus on livestream categories primarily related to entertainment, including genres like singing, dancing, and chatting. The distribution of these vertical domains is shown in the lower left part of Fig.~\ref{fig:visu1}.

\paragraph{Video data filtering} To ensure video diversity, we filter out static and simple videos with minimal frame changes. We employ a proprietary model, Seed1.5-VL, to score each video's spatiotemporal complexity on a scale from 1 to 10, and filter out those scoring below 3. We also exclude videos that focus on web games or e-commerce. After filtering over 30,000 videos, we retain 5,245 videos with durations ranging from 20 seconds to 60 minutes, which form the basis of our benchmark.

\begin{figure}
    \centering
    \includegraphics[width=0.48\textwidth]{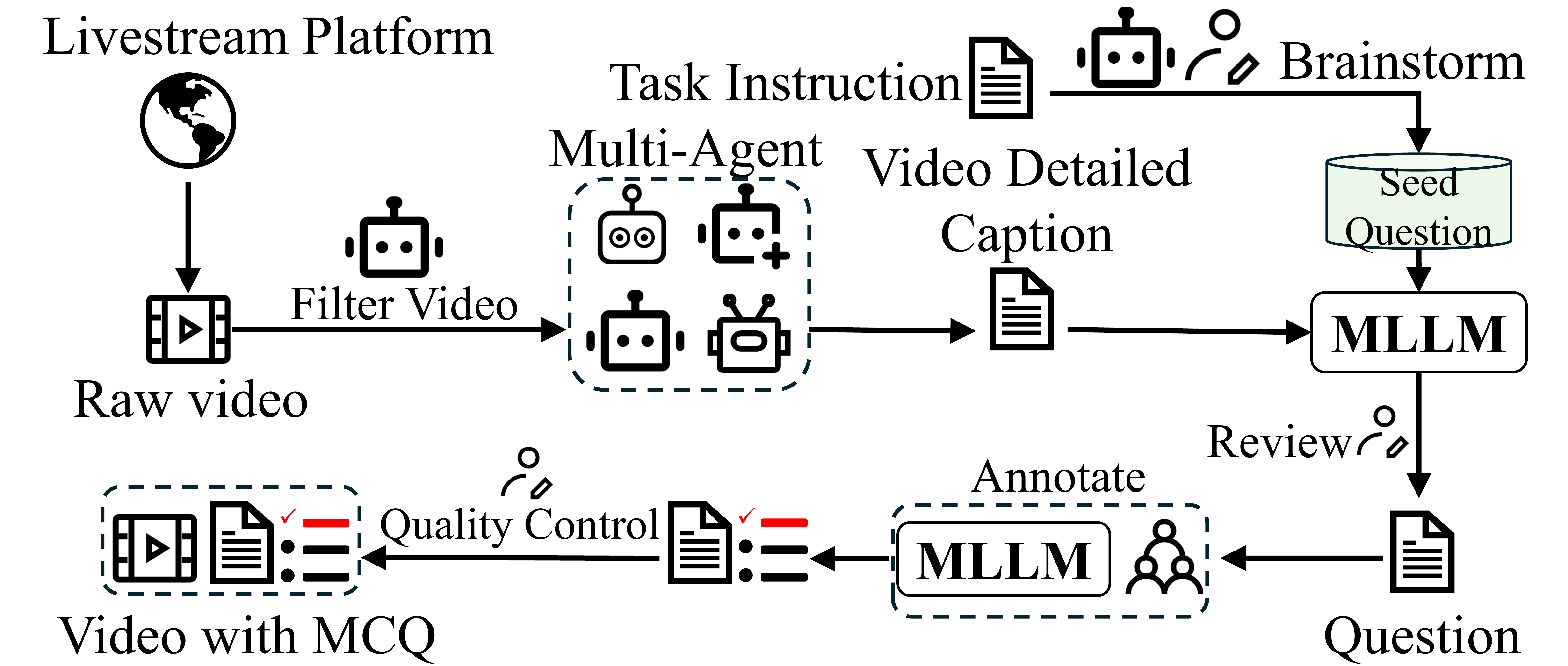}
    \caption{Dataset generation pipeline.}
    \vspace{-5mm}
    \label{fig:pipe}
\end{figure}

\subsection{Multi-Agent Seed-guided Video QA Generation Pipeline}
\label{detail_pipe}
Constructing a video benchmark requires detailed annotation of the video. At the same time, designing and answering questions based on videos demands careful handling of temporal information, which makes the annotation process more difficult and costly. Previous works~\cite{fu2025videomme,wu2024longvideobench}  rely entirely on manual annotation to design questions and answers, involving high labor costs. Recent fully automated efforts, such as~\cite{han2025videoespresso}, rely solely on a single model for video captioning and then use GPT-4o to generate questions and answers. However, this method introduces too many biases from the captioning model, and the result questions and options tend to be lengthy, influenced by the preferences of the language model.

To address these issues, we propose a novel video QA generation pipeline as shown in Fig.~\ref{fig:pipe}, characterized by multi-agent annotation, seed-based question-posing, and human-in-the-loop at multiple stages. To mitigate the impact of model bias in video descriptions, we build a multi-agent system composed of several large multimodal models with large parameter sizes, including LLaVA-Video, Qwen2.5-VL, Intern3VL, and Seed1.5-VL. Based on their unique characteristics and capabilities, we develop specialized instructions for each agent, ensuring they generate only the content necessary for their specific tasks. Through the collaboration of different experts, the resulting detailed video description is no longer limited by the capabilities of a single model, but contains richer and more comprehensive content.

We design 24 tasks grouped into 5 categories: 4 coarse-grained perception tasks, 6 fine-grained perception tasks, 3 knowledge-based reasoning tasks, 4 general reasoning tasks, and 7 livestream-specific tasks. To ensure controllability and high-quality QA generation, we introduce a seed question–driven framework that generates questions based on detailed video descriptions. Specifically, we first define task-specific instructions and employ a proprietary model to extract and summarize question patterns from previous work, generating candidate seed questions for each task.  Human annotators then review the questions to remove unreasonable or overly simple questions, revising them as needed to form a curated seed question library. Leveraging this library and detailed video descriptions from the multi-agent pipeline, the proprietary model generates candidate questions tailored to specific tasks.  Finally, annotators review both the videos and the generated questions, filtering or refining those that are ambiguous, overly simple, or irrelevant.
For the video question set of all tasks after quality control, we employ a proprietary model and human annotators to generate answers and propose alternative distractors, forming the initial multiple-choice questions. Subsequently, for each question, annotators are required to carefully review the corresponding video content to verify the correctness of the answer and remove or refine any misleading or inappropriate distractors.

\subsection{Data Analysis}
\begin{figure}
    \centering
    \includegraphics[width=\linewidth]{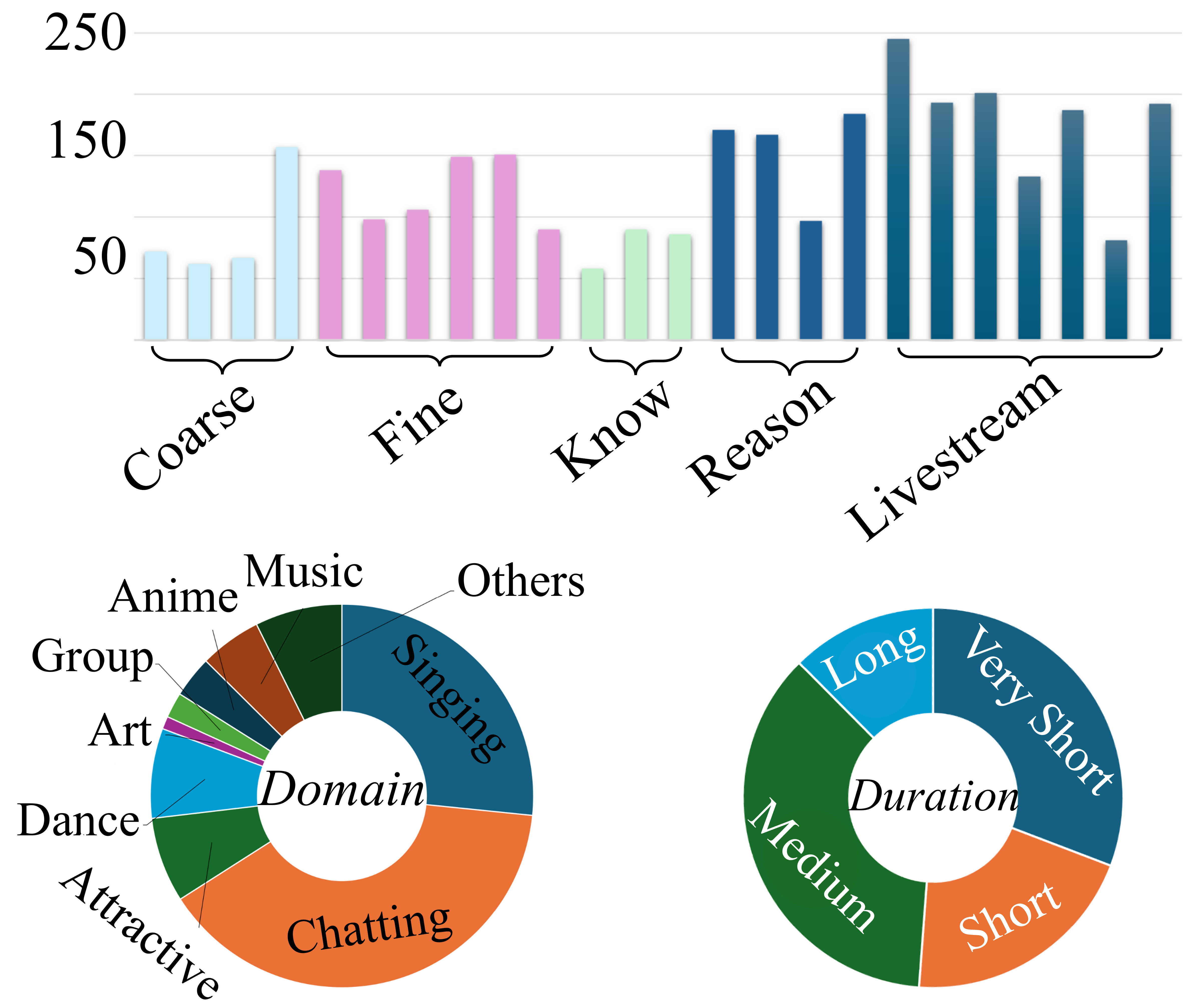}
    \caption{The statistical analysis of our LiViBench.}
    \label{fig:visu1}
\end{figure}

This section presents a detailed data analysis. The dataset includes 3,168 videos, each with audio, comments, and ASR information. As shown in Fig.\ref{fig:visu1}, the videos mainly belong to entertainment-related livestream categories. Their durations range from 14 seconds to 33 minutes and are grouped into four categories: very short, short, medium, and long. The duration and task distribution are also shown in ~Fig.\ref{fig:visu1}.

\begin{figure}[t]
    \centering
    \includegraphics[width=\linewidth]{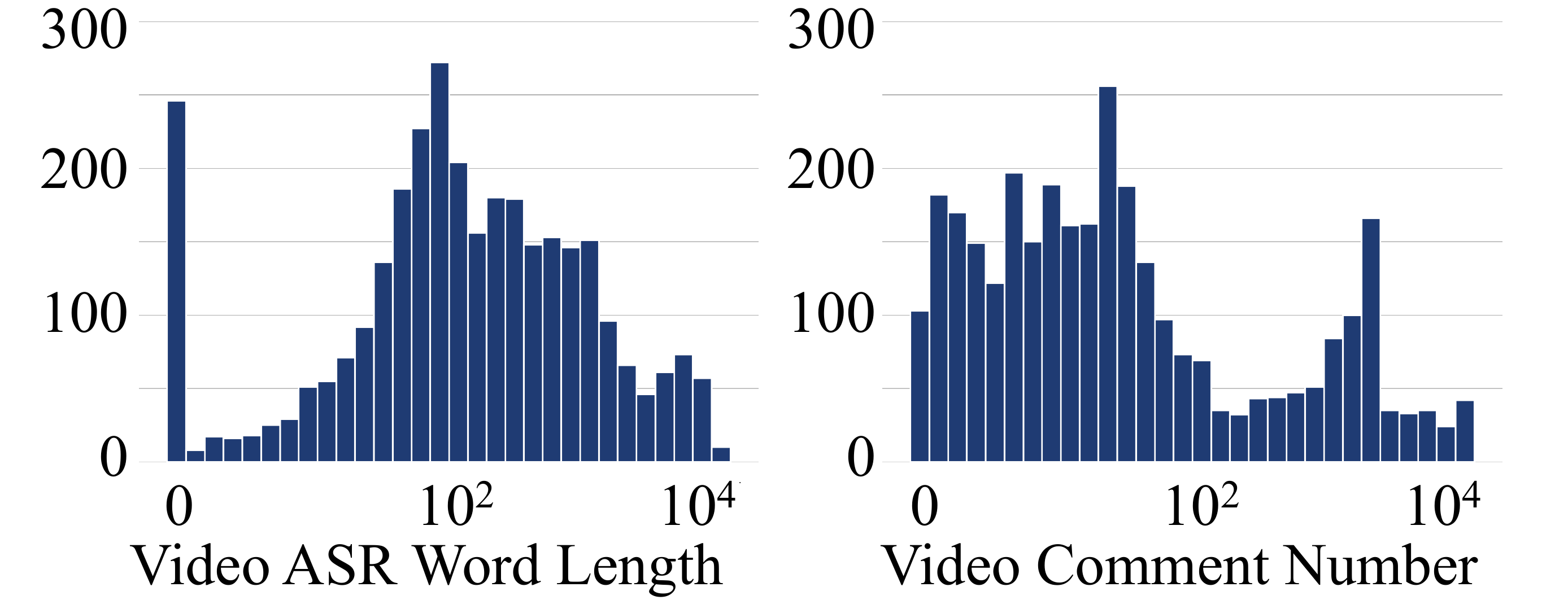}
    \caption{The ASR and comment distribution.}
    \label{fig:dis}
\end{figure}

\begin{figure}
    \centering
    \includegraphics[width=\linewidth]{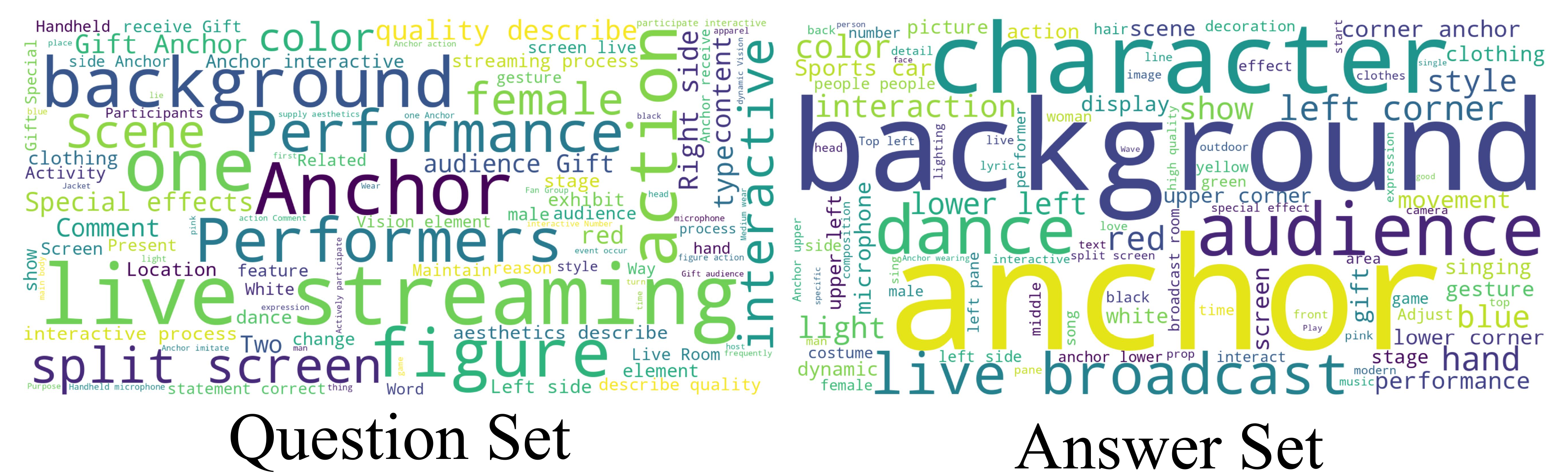}
    \caption{The word clouds of our LiViBench.}
    \label{fig:cloud}
\end{figure}

Due to the real-time interactive nature of livestream videos, they are often accompanied by numerous real-time comments and automatic speech recognition (ASR) transcripts. We analyze the distribution of ASR and comment counts within the videos, as shown in Fig.~\ref{fig:dis}. The dataset contains approximately 1.45 million comments, with an average length of 12.15 Chinese characters. To illustrate the characteristics of livestream videos, we generate word clouds based on the question and option sets, shown in Fig.~\ref{fig:cloud}. In the question set, terms like “performance,” “action,” and “interaction” appear frequently, while words such as “anchor” and “audience” dominate the option set. This distribution reflects the benchmark’s focus on livestream-specific features, particularly those related to performance dynamics and interactive behavior.

\section{LiVi-LLM}

To enhance the ability of open-source MLLMs to understand interactive livestream videos, we construct an instruction-tuning dataset enriched with interactive knowledge. Based on a carefully designed two-stage instruction tuning strategy and the efficient use of comments through a Video-to-Comment Retrieval (VCR) module in Fig.~\ref{fig:train}, we develop LiVi-LLM-7B, a high-performance and efficient omnimodal model for livestream video understanding. The following sections detail the model’s training and inference processes.

\subsection{Instruction Tuning}
\begin{figure*}
    \centering
    \includegraphics[width=0.8\textwidth]{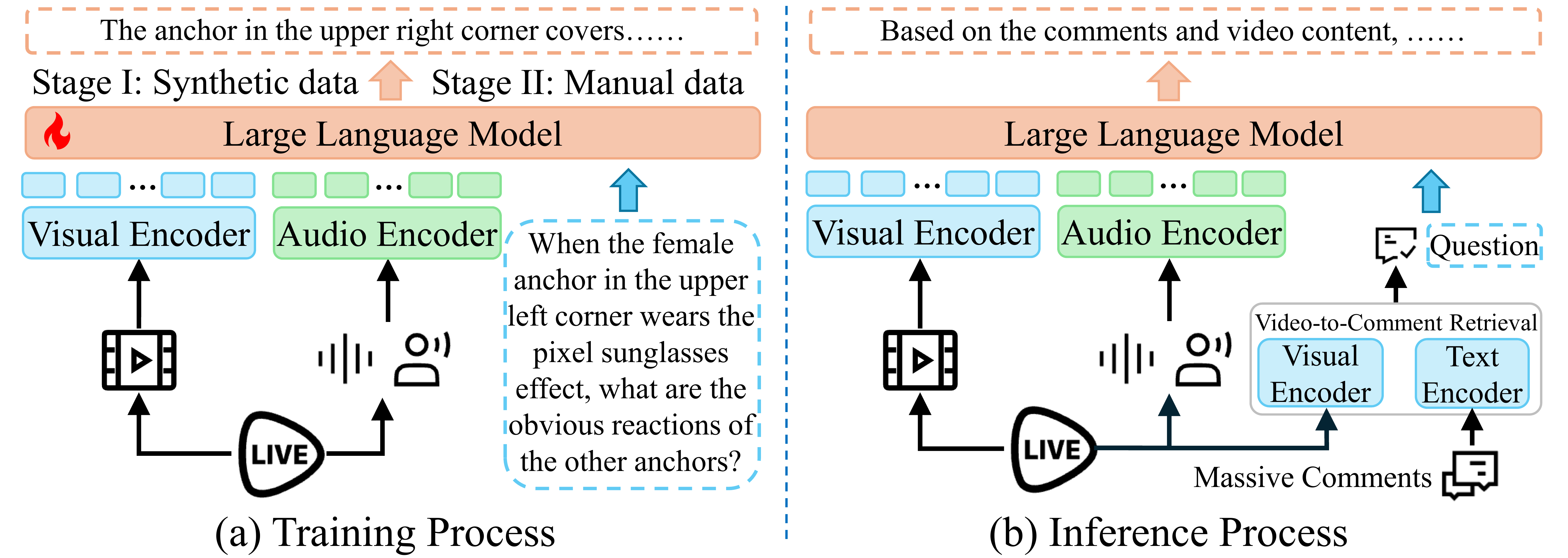}
    \caption{LiVi-LLM architecture. (a) Training process: In the first stage, the model is aligned to the interactive video domain using synthetic data; in the second stage, it undergoes fine-grained tuning with manual data. (b) Inference process: The model integrates a video-to-comment retrieval module to fully use omni-modality to enhance the comprehensive understanding. }
    \label{fig:train}
\end{figure*}

The left part of Fig.~\ref{fig:train} illustrates the architecture and training process of the model. Given a livestream video, we extract the video frames and audio from it, respectively. For continuous video frames, we use Qwen2.5-VL's visual encoder for spatiotemporal encoding and convert them into video tokens. For audio streams, we use Qwen2-Audio to encode them into audio tokens. To fuse the video and audio representations, a transformer decoder is used to aggregate the features. The fused tokens are then input into the large language model along with the query's text tokens. The parameters of the model are initialized from Qwen-2.5-Omni.

The data generation pipeline proposed in Sec.\ref{detail_pipe} can be directly extended to the construction of instruction fine-tuning data in the livestream field. We collect over 100,000 livestream videos, each ranging from 1 to 5 minutes in length. To ensure content richness, we use a multimodal large language model to analyze the number of distinct scenes in each video and retain only those containing more than one scene. For each selected video, agents consisting of one or more models generate 1 to 3 questions along with corresponding answers. To improve the accuracy of instruction tuning data while balancing annotation costs, we sample a portion of the data for human review and annotation. These higher-quality samples are used for fine-grained tuning of the model. In total, we obtain 37,953 synthetic samples without human review and 11,180 manually refined samples.

During instruction tuning, we adopt a carefully designed two-stage training strategy. In the first stage, the model is fine-tuned on synthetic data without manual review to align the model to the interactive video domain. To balance domain specificity with generalization, we also incorporate general video data~\cite{zhang2024llavavideo}, helping the model retain general video understanding capabilities. In the second stage, we perform fine-grained tuning using manually annotated data to further enhance the model’s accuracy and robustness on video understanding tasks.

\subsection{Inference with Video-to-Comment Retrieval}
Instruction tuning on the specific interactive video domain can enhance the understanding ability of MLLMs for livestream videos. But the core difference between livestream videos and general videos is real-time interaction, such as real-time user comments during live streaming. However, the massive amount of real-time comments poses a huge challenge to the model's context and information extraction capabilities. To this end, this paper proposes a Video-to-Comment Retrieval (VCR) module that aims to obtain more relevant comments from massive comments. For efficient retrieval, we uniformly sample video frames from a video, and use Chinese-CLIP~\cite{chinese-clip} to obtain embeddings for each frame. We use a text encoder to encode all comments into text embeddings. By calculating the similarity between frame embeddings and text embeddings, we can obtain the top-k relevant comments corresponding to each frame. All retrieved relevant comments are sorted in chronological order, and together with the question as text context, they are input into the fine-tuned model.

\section{Experiment}

\begin{figure*}
    \centering
    \includegraphics[width=0.75\linewidth]{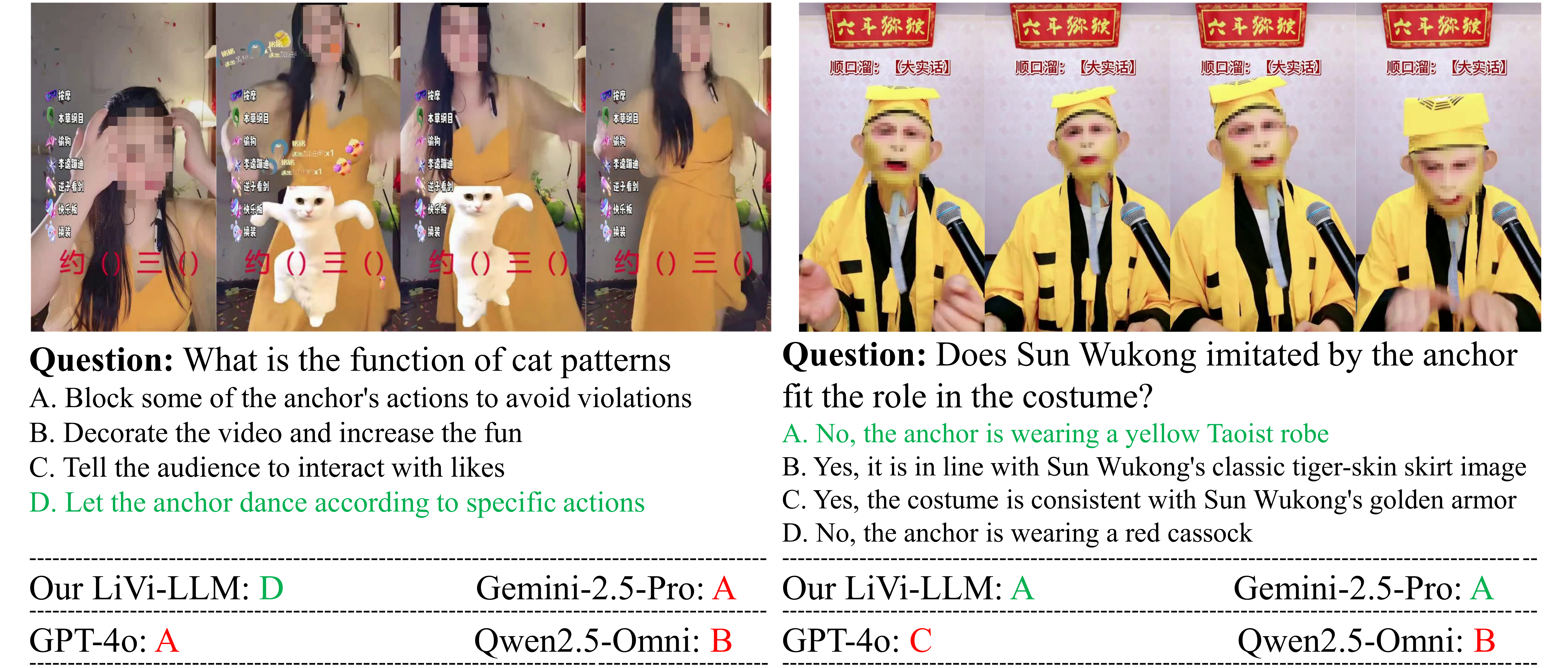}
    \caption{Qualitative examples of multi-choice questions from our LiViBench. The correct options are marked in green.}
    \label{fig:quality}
\end{figure*}

\subsection{Settings}
The proposed benchmark includes a total of 3175 Video QA. We evaluate 24 models and use the most suitable number of frames for each model for reasoning while ensuring the context does not overflow. The benchmark comprises 24 subtasks, categorized into five evaluation groups: 4 coarse-grained general perception tasks (Coarse), 6 fine-grained general perception tasks (Fine), 3 knowledge-based question answering tasks (Know), 4 general reasoning tasks (Reason), and 7 livestream-specific tasks (Livestream). For evaluation, we report the average score of all models within each category and the overall score of the entire benchmark. More details refer to the supplementary material.

\begin{table*}[]
\vspace{-3mm}
    \centering
    \small
    \begin{tabular}{l|cccccc}
    \toprule
    \textbf{Model} & \textbf{Overall} & {\textbf{Coarse}} & {\textbf{Fine}}  & {\textbf{Know}} & {\textbf{Reason}} & {\textbf{Livestream}} \\ 
    
    \midrule
    \multicolumn{7}{c}{\textbf{Proprietary MLLMs}} \\

    Gemini 2.5 Flash~\cite{comanici2025gemini25} & 53.0 & 63.6 & 62.9 & 56.4 &  51.5 & 43.9 \\

    Gemini 2.5 Pro~\cite{comanici2025gemini25} & 56.1 & 65.0 & 68.4 &58.1 &51.3 &48.2   \\

    GPT-4o~\cite{hurst2024gpt4o} & 56.3 & 67.0 & 66.5 & 57.6 & 55.2 & 47.4 \\
    
    Seed1.5-VL~\cite{guo2025seed15vl}    & \textbf{66.2} & 70.9 & 71.4 & \textbf{68.8} & \textbf{70.7} & \textbf{59.1}   \\
    Doubao-Seed-1.6~\cite{seed16} & 64.9 & \textbf{72.9} & \textbf{73.2} &60.2 &68.4 &56.8  \\

    \midrule
    \multicolumn{7}{c}{\textbf{Open-Source MLLMs}} \\
    
    {LLaVA-Video-72B~\cite{zhang2024llavavideo}} & {60.0} & {65.3} & {70.2} & {63.6} & {63.8} & {49.8} \\

    {Qwen2.5-VL-32B~\cite{bai2025qwen25vl}} & {59.4} & {73.1} & {69.1} & {57.6} & {61.3} & {49.1} \\
    
    {Qwen2.5-VL-72B~\cite{bai2025qwen25vl}} & {62.3} & {73.4} & {72.4} & {61.9} & {64.6} & {52.0} \\

    {InternVL3-14B~\cite{zhu2025internvl3}} & {62.7} & {71.7} &{68.9} & {65.3} & {67.0} & {53.7}\\  

    {InternVL3-38B~\cite{zhu2025internvl3}} & 64.1 & 70.9 & 72.6 & 66.6 & 68.3 & 54.5 \\
    
    {InternVL3-78B~\cite{zhu2025internvl3}} & {64.4} & {72.0} &{69.8} & {65.8} & {69.3} & {56.3}\\  

    \midrule
    
    LLaVA-NeXT-Video~\cite{zhang2024llavanextvideo} & 37.5 & 40.7 & 40.5 &44.8 &32.3 & 36.1 \\
    InternVL2-8B~\cite{chen2024internvl2} & 49.8 & 59.7  & 59.2 &53.8 &53.6 &38.7\\
    MiniCPM-v-26~\cite{yao2024minicpm} & 50.5 & 58.9 & 59.9 &58.5 &54.2 &39.1 \\
    LLaVA-Video-7B~\cite{zhang2024llavavideo} & 52.6 & 60.0 & 59.0 &55.1 &58.3 &43.5 \\
    
    NVILA-8B-Video~\cite{liu2025nvila} & 53.3 & 61.1 & 56.2 & 58.1  & 56.8 &46.6\\
    Video-LLaMA3-8B~\cite{zhang2025videollama3} & 54.1 & 60.0 & 62.1 &61.9 &58.8 &43.7 \\
    Keye-VL-8B-Preview~\cite{team2025keye} & 55.2 & 68.7 & 65.4 & 51.7 &52.6 &47.2 \\
    MiniCPM-o-26$^\dagger$~\cite{yao2024minicpm} & 56.0 & 65.3 & 62.1 &61.9 &59.9 &46.5\\
    InternVL2.5-8B~\cite{chen2024internvl25} & 56.6 & 68.1 & 62.9 &58.5 &59.9 &47.5\\
    Qwen2.5-VL-7B~\cite{bai2025qwen25vl} & 58.3 & 65.5 &69.2 &57.6 &59.7 &49.1 \\
    
    InternVL3-8B~\cite{zhu2025internvl3} & 59.8 & 68.4 &66.9 & 58.5 &63.0 &51.7 \\

    InternVL3-9B~\cite{zhu2025internvl3} & 60.0 & 67.3 &67.8 &59.4 &63.0 &51.7 \\
    
    Qwen2.5-Omni-7B$^\dagger$~\cite{xu2025qwen25omni} & 60.3 & 68.1 &68.5 &59.4 &60.7 &53.1\\

    \midrule

    LiVi-LLM-7B$^\dagger$ (Ours) & \textbf{64.4} & \textbf{70.1} & \textbf{68.7} & \textbf{62.8} & \textbf{63.6} & \textbf{60.9} \\

    \bottomrule

    \end{tabular}
    \caption{LiViBench evaluation results across all categories. $^\dagger$ indicates omnimodal models.}
    \label{tab:main_result}
\end{table*}

\subsection{Main Results on LiViBench}
The comprehensive evaluation results for LiViBench are presented in Tab.~\ref{tab:main_result}. The results indicate that general perception tasks (e.g., Coarse and Fine) consistently outperform other task types in terms of accuracy, for both proprietary and open-source models. The livestream-specific tasks are the most challenging for all models. Among proprietary models, Gemini-2.5-Pro and GPT-4o show limited performance on this benchmark. In contrast, Doubao-Seed-1.6 and Seed1.5-VL show promising results, with Seed1.5-VL attaining the highest overall score of 66.2\%. Doubao-Seed-1.6 performs best on Coarse and Fine tasks, while Seed1.5-VL leads in tasks including knowledge, reasoning, and livestream-specific, which indicates that the model has superior expert knowledge and reasoning ability in the livestream video domain. Among all open-source models, our model LiVi-LLM-7B significantly outperforms other large-scale open-source models with up to 72B parameters, including Qwen2.5-VL-72B. It also surpasses proprietary models such as GPT-4o and Gemini-2.5-Pro. Notably, LiVi-LLM achieves the best overall accuracy of 64.4\%, matching the top-performing InternVL3-78B model. To showcase the unique features of the benchmark, we provide qualitative examples in Fig.~\ref{fig:quality}. Compared to general proprietary models, our model demonstrates enhanced understanding and reasoning abilities in interactive video scenarios.

\subsection{Results on General Video Benchmarks}
To demonstrate the comprehensive capabilities of our model, we conduct experiments on various general video benchmarks, including Video-MME~\cite{fu2025videomme}, MLVU~\cite{zhou2024mlvu}, LongVideoBench (LongVB)~\cite{wu2024longvideobench}, and VideoEval-Pro~\cite{ma2025videoeval}. In Tab.~\ref{tab:general}, compared with the state-of-the-art models of similar parameter size, our model demonstrates promising results across all benchmarks, including the best scores on all tasks of Video-MME and VideoEval-Pro. These results indicate that our model not only excels in interactive live video understanding but also exhibits strong generalization capabilities.

\begin{table*}[]
    \centering
    \small
    \begin{tabular}{l|C{4mm}C{4mm}C{4mm}C{4mm}C{4mm}C{4mm}C{4mm}C{4mm}C{4mm}C{4mm}C{4mm}C{4mm}C{4mm}C{4mm}C{4mm}C{4mm}C{4mm}C{4mm}C{4mm}}
    \toprule
    \textbf{Model}  & \multicolumn{3}{c}{\textbf{Overall}} & \multicolumn{3}{c}{\textbf{Coarse}} & \multicolumn{3}{c}{\textbf{Fine}}  & \multicolumn{3}{c}{\textbf{Know}} & \multicolumn{3}{c}{\textbf{Reason}} & \multicolumn{3}{c}{\textbf{Livestream}} \\ 
    & V & +A & +S & V & +A & +S & V & +A & +S & V & +A & +S & V & +A & +S & V & +A & +S \\
    \midrule
LLaVA-Video-7B & 52.6 & NA & 55.4↑ & 60.0 & NA & 60.6↑ & 59.0 & NA & 61.0↑ &55.1 & NA & 58.1↑ &58.3 & NA & 58.8↑ &43.5 & NA & 48.4↑  \\

MiniCPM-o-26 & 56.0 & 54.7 & 57.9↑  & 65.3 & 65.3 & 67.0↑ & 62.1 & 61.2 & 61.8 & 61.9 & 58.5 & 62.8↑ & 59.9 & 51.6 & 59.7 & 46.5 & 48.5↑ & 51.2↑ \\

Qwen2.5-Omni-7B & 57.8 & 60.3↑ & 60.2↑ & 66.2 & 68.1↑ & 67.5↑ & 67.4 & 68.5↑ & 66.9 & 57.2 & 59.4↑ & 59.4↑ & 62.6 & 60.7 & 63.1↑ & 47.4 & 53.1↑ & 52.8↑  \\

Qwen2.5-VL-7B & 58.3 & NA & 59.8↑ & 65.5 & NA & 67.0↑ & 69.2 & NA & 68.4 & 57.6 & NA & 60.2↑ & 59.7& NA & 59.4 & 49.1 & NA & 52.7↑ \\

InternVL3-8B & 59.8 & NA & 61.4↑ & 68.4 & NA & 70.6↑ & 66.9 & NA & 65.3 & 58.5 & NA & 59.4↑ & 63.0 & NA & 63.3↑ & 51.7 & NA & 55.8↑ \\

\midrule
LiVi-LLM-7B & 61.4 & \textbf{63.9}↑ & 63.4↑ & 70.6 & \textbf{70.9}↑ & 68.7 & \textbf{69.1} & 68.6 & 68.7 & 63.2 & \textbf{65.3}↑ & 62.8 & 61.7 & \textbf{64.2}↑ & 62.1↑ & 53.6 & 58.7↑ & \textbf{59.6}↑\\

    \bottomrule
\end{tabular}
    \caption{Analysis of the impact of audio and speech. V: Video, A: Audio, S: Speech. We use ASR to represent speech modality.}
    \label{tab:asr}
\end{table*}

\begin{table*}[]
    \centering
    \small
    \begin{tabular}{l|ccccccc}
    \toprule
    \textbf{Model} &  \multicolumn{4}{c}{\textbf{Video-MME}}  & {\textbf{MLVU}} & {\textbf{LongVB}} & {\textbf{VideoEval-Pro}} \\ 
     & Short & Med & Long & Overall & M-Avg & val total & MCQ \\
    \midrule
    \multicolumn{8}{c}{\textbf{Proprietary MLLMs}} \\

    GPT-4o~\cite{hurst2024gpt4o} & 80.0 & 70.3 & 65.3 & 71.9 & 64.6 & 66.7 & 59.5 \\
    Seed1.5-VL~\cite{guo2025seed15vl} & - & - & - & \textbf{77.9} & \textbf{82.1} & \textbf{74.0} & \textbf{66.6} \\
    Doubao-Seed-1.6~\cite{seed16} & 84.1 & 75.3 & 70.1 & 76.5 &76.0 & 71.3 & 62.4 \\
    \midrule
    \multicolumn{8}{c}{\textbf{Open-Source MLLMs}} \\
    InternVL2-8B~\cite{chen2024internvl2} & 68.0 & 52.0 & 48.9 & 56.3 & 56.3 & 54.6 & 39.9  \\
    
    Qwen2.5-VL-7B~\cite{bai2025qwen25vl} & 75.9 & 66.8 & 54.1 & 65.6 & 65.1 & \textbf{61.0} & 46.9 \\

    InternVL2.5-8B~\cite{chen2024internvl25} & 75.3 & 61.5 & 55.8 & 64.2 & 68.9 & 60.0 & 45.5  \\
    
    InternVL3-8B~\cite{zhu2025internvl3} & 77.5 & 67.3 & 54.1 &  66.3 & \textbf{71.4} & 58.8 & 48.4 \\

    Qwen2.5-Omni-7B~\cite{xu2025qwen25omni} & 78.1 & 67.4 & \textbf{57.6} & 67.7 & 70.0 & 58.7 & 48.9 \\

    \midrule
    LiVi-LLM-7B (Ours) & \textbf{79.6} & \textbf{69.4} & 56.6 & \textbf{68.5} & 70.5 & 59.6 & \textbf{50.5} \\
    
    \bottomrule
\end{tabular}
    \caption{Evaluation results on general video benchmarks.}
    \label{tab:general}
\end{table*}

\begin{table*}[]
\vspace{-1mm}
    \centering
    \small
    \begin{tabular}{l|C{3mm}C{4mm}C{6.5mm}|C{3mm}C{4mm}C{6.5mm}|C{3mm}C{4mm}C{6.5mm}|C{3mm}C{4mm}C{6.5mm}|C{3mm}C{4mm}C{6.5mm}}
    \toprule
    \textbf{Level} &  \multicolumn{3}{c}{\textbf{InternVL3-8B}}  & \multicolumn{3}{c}{\textbf{Qwen2.5-VL-7B}}  & \multicolumn{3}{c}{\textbf{LLaVA-Video-7B}}  & \multicolumn{3}{c}{\textbf{Qwen2.5-Omni-7B}}  &   \multicolumn{3}{c}{\textbf{LiVi-LLM-7B}} \\ 
     & V &+Raw  &+VCR & V & +Raw & +VCR &V & +Raw & +VCR & V & +Raw & +VCR & V & +Raw & +VCR \\
    \midrule
    $[0, 20)$ & 60.6 & 59.3 & 58.8 & 59.9 & 59.3 & 59.0 & 52.8  & 54.1↑ & 54.3↑ &60.4 & 60.8↑ & 61.3↑ & 63.7 & 63.6 & \textbf{63.8}↑  \\
    $[20, 100)$ & 56.6 & 60.7↑ & 61.6↑ & 53.5 & 55.5↑ & 56.3↑ & 47.9  &  51.1↑ & 53.0↑ & 60.3  & 58.7 & 57.7 & 64.2 & \textbf{65.7}↑ & 65.6↑       \\
    $[100, 1\text{k})$  & 62.0  & 61.0  & 61.2 & 62.6 & 56.9  & 58.8 & 56.4 &  57.2↑  & 57.7↑ & 62.3 & 62.0 & 64.5↑ & 64.5 &  64.5 &   \textbf{66.1}↑      \\
    $[1\text{k}, \infty)$ & 60.6  & 59.9 & 63.4↑ & 57.5 & 49.0 & 54.9 & 57.1 & 29.5 & 58.8↑ & 58.2 & 54.7 & 59.7↑  & \textbf{63.8} & 55.3 &  63.0     \\

     Overall &  59.8 & 60.0↑ & 60.4↑ & 58.3  & 56.6 & 57.7 &  52.6 & 50.2 & 55.0↑ & 60.3  & 59.5 & 60.5↑  & 63.9 &  63.0 & \textbf{64.4}↑ \\

    \bottomrule
\end{tabular}
\vspace{-2mm}
    \caption{Analysis video comments impact. Raw: using raw comments. VCR: using Video-to-Comment Retrieval module.}
    \label{tab:comment}
    \vspace{-3mm}
\end{table*}

\begin{table}[]
    \centering
    \small
    \begin{tabular}{ll|cccc}
    \toprule
        \textbf{Stage \uppercase\expandafter{\romannumeral1}} & \textbf{Stage \uppercase\expandafter{\romannumeral2}} & \textbf{LiViBench} &  \multicolumn{3}{c}{\textbf{Video-MME}}  \\ 
    & &  Overall & Short & Med & Long \\
    \midrule
    Baseline & - & 60.3 & 78.1 & 67.4 & 57.6 \\
    \midrule
     Ours &  \ding{55}  & 62.9 & \underline{80.0} & \underline{69.2} & \underline{57.7}  \\
     Ours &  Ours  & \textbf{63.9} & 79.6 & \textbf{69.4} & 56.6 \\
     LV+Ours &  \ding{55}   & 62.2 & \textbf{80.6} & \textbf{69.4} & \textbf{57.8}  \\
     
     LV+Ours &  Ours  & \underline{63.1}  & \textbf{80.6} & \textbf{69.4} & 57.3 \\
     \bottomrule
    \end{tabular}
    \caption{Ablation study of the data used in different training stages. LV indicates data from LLaVA-Video-178k.}
    \label{tab:data_domain}
    \vspace{-5mm}
\end{table}

\subsection{Analysis}

\paragraph{Audio impact analysis}
The results of the impact of audio for omnimodal models are in Tab.~\ref{tab:asr}, including MiniCPM-o-26, Qwen2.5-Omni and LiVi-LLM-7B. MiniCPM-o-26 shows a performance drop across the first four categories, likely due to its limited ability to process audio. In comparison, both Qwen2.5-Omni and our model show notable improvement in most categories. We observe that all 3 models showed significant improvements in the livestream-specific category. This suggests that in the interactive video domain, audio plays a crucial role in improving video understanding.

\paragraph{Speech impact analysis} The results of the impact of speech modality on model evaluation are also shown in Tab.~\ref{tab:asr}. We use ASR to represent the speech modality. All models perform better in most categories and obtain an overall performance improvement. Using speech in fine-grained and reasoning tasks sometimes degrades model performance, suggesting that speech can sometimes be noisy and affect the understanding of video details. Comparing audio and speech modalities, we find that audio modality is more helpful. These findings highlight the importance of effectively leveraging modality information in interactive video understanding and may inspire future work.

\paragraph{Video comment impact analysis}
In order to analyze the processing capabilities of different models for comments, we divide all the comments in the benchmark into 4 levels according to comment number, including [0, 20), [20, 100), [100, 1k), [1k, $\infty$), as shown in Tab.~\ref{tab:comment}.
Using raw comments degrades the performance of most models, but InternVL3 is an exception since it uses far fewer frames and thus has more context space. To mitigate the adverse effect of massive comments, we propose VCR module. This module retrieves key comments, reduces the adverse effects of massive comments, and brings better performance than video-only.

\subsection{Ablation Study}

\paragraph{Impact of training data domain}
We test the data domain used in different stages in Tab.~\ref{tab:data_domain}. We test different data used in the first stage. We can see that the first stage alignment training using only our synthetic data achieved better results on LiViBench. When general data~\cite{zhang2024llavavideo} is added, performance on LiViBench decreases but improves across all three categories of Video-MME. In the second stage, further fine-tuning using our manual data continuously improves the performance of LiViBench while maintaining good generalization. This shows that our training data is effective in improving the performance on both LiViBench and general benchmarks. However, there is still a trade-off between interactive understanding tasks and general tasks. These findings indicate that interactive video data contributes meaningfully in both training stages, with the second stage playing a key role in strengthening the model’s interactive livestream knowledge.

\section{Conclusion}
This paper introduces LiViBench, the first omnimodal benchmark for interactive livestream video understanding. To construct LiViBench, we develop a standardized semi-automatic workflow for efficient annotation and data quality. We incorporate audio, speech, and real-time comments to build an omnimodal benchmark and perform comprehensive evaluation and analysis. In addition, we carefully build LiVi-LLM-7B, a video model enhanced with interactive knowledge that outperforms open-source models with up to 72B parameters, establishing a strong baseline and a new foundation for future research in interactive video understanding.

\section{Acknowledgments}
The study was funded by the Shenzhen Science and Technology Program (KQTD20240729102051063), the National Natural Science Foundation of China under contracts No. 62422602, No. 62372010, No. 62425101, No. 62332002, No. 62372010, and No. 62206281.

\bibliography{aaai2026}

@String(CVPR= {IEEE Conf. Comput. Vis. Pattern Recog.})

@String(AAAI = {AAAI})

@String(CVPR  = {CVPR})

@article{wang2025leanpo,
  title={LeanPO: Lean Preference Optimization for Likelihood Alignment in Video-LLMs},
  author={Wang, Xiaodong and Huang, Jinfa and Yuan, Li and Peng, Peixi},
  journal={arXiv preprint arXiv:2506.05260},
  year={2025}
}

@misc{wang2025open,
  title={Open-r1-video},
  author={Wang, Xiaodong and Peng, Peixi},
  year={2025}
}

@article{chinese-clip,
  title={Chinese CLIP: Contrastive Vision-Language Pretraining in Chinese},
  author={Yang, An and Pan, Junshu and Lin, Junyang and Men, Rui and Zhang, Yichang and Zhou, Jingren and Zhou, Chang},
  journal={arXiv preprint arXiv:2211.01335},
  year={2022}
}

@article{hurst2024gpt4o,
  title={Gpt-4o system card},
  author={Hurst, Aaron and Lerer, Adam and Goucher, Adam P and Perelman, Adam and Ramesh, Aditya and Clark, Aidan and Ostrow, AJ and Welihinda, Akila and Hayes, Alan and Radford, Alec and others},
  journal={arXiv preprint arXiv:2410.21276
        
        
        
        
        
        
        
        
        
        },
  year={2024}
}

@article{bai2025qwen25vl,
  title={Qwen2. 5-vl technical report},
  author={Bai, Shuai and Chen, Keqin and Liu, Xuejing and Wang, Jialin and Ge, Wenbin and Song, Sibo and Dang, Kai and Wang, Peng and Wang, Shijie and Tang, Jun and others},
  journal={arXiv preprint arXiv:2502.13923
        
        
        
        
        
        
        
        
        
        },
  year={2025}
}

@article{zhu2025internvl3,
  title={Internvl3: Exploring advanced training and test-time recipes for open-source multimodal models},
  author={Zhu, Jinguo and Wang, Weiyun and Chen, Zhe and Liu, Zhaoyang and Ye, Shenglong and Gu, Lixin and Tian, Hao and Duan, Yuchen and Su, Weijie and Shao, Jie and others},
  journal={arXiv preprint arXiv:2504.10479
        
        
        
        
        
        
        
        
        
        },
  year={2025}
}

@inproceedings{msrvtt,
  title={Msr-vtt: A large video description dataset for bridging video and language},
  author={Xu, Jun and Mei, Tao and Yao, Ting and Rui, Yong},
  booktitle={Proceedings of the CVPR},
  pages={5288--5296},
  year={2016}
}

@inproceedings{yu2019activitynet,
  title={Activitynet-qa: A dataset for understanding complex web videos via question answering},
  author={Yu, Zhou and Xu, Dejing and Yu, Jun and Yu, Ting and Zhao, Zhou and Zhuang, Yueting and Tao, Dacheng},
  booktitle={Proceedings of the AAAI},
  volume={33},
  pages={9127--9134},
  year={2019}
}

@inproceedings{li2024mvbench,
  title={Mvbench: A comprehensive multi-modal video understanding benchmark},
  author={Li, Kunchang and Wang, Yali and He, Yinan and Li, Yizhuo and Wang, Yi and Liu, Yi and Wang, Zun and Xu, Jilan and Chen, Guo and Luo, Ping and others},
  booktitle={Proceedings of the CVPR},
  pages={22195--22206},
  year={2024}
}

@article{fang2024mmbench,
  title={Mmbench-video: A long-form multi-shot benchmark for holistic video understanding},
  author={Fang, Xinyu and Mao, Kangrui and Duan, Haodong and Zhao, Xiangyu and Li, Yining and Lin, Dahua and Chen, Kai},
  journal={NeurIPS},
  volume={37},
  pages={89098--89124},
  year={2024}
}

@inproceedings{hong2025motionbench,
  title={Motionbench: Benchmarking and improving fine-grained video motion understanding for vision language models},
  author={Hong, Wenyi and Cheng, Yean and Yang, Zhuoyi and Wang, Weihan and Wang, Lefan and Gu, Xiaotao and Huang, Shiyu and Dong, Yuxiao and Tang, Jie},
  booktitle={Proceedings of the CVPR},
  pages={8450--8460},
  year={2025}
}

@article{liu2024tempcompass,
  title={Tempcompass: Do video llms really understand videos?},
  author={Liu, Yuanxin and Li, Shicheng and Liu, Yi and Wang, Yuxiang and Ren, Shuhuai and Li, Lei and Chen, Sishuo and Sun, Xu and Hou, Lu},
  journal={arXiv preprint arXiv:2403.00476},
  year={2024}
}

@article{shangguan2024tomato,
  title={Tomato: Assessing visual temporal reasoning capabilities in multimodal foundation models},
  author={Shangguan, Ziyao and Li, Chuhan and Ding, Yuxuan and Zheng, Yanan and Zhao, Yilun and Fitzgerald, Tesca and Cohan, Arman},
  journal={arXiv preprint arXiv:2410.23266},
  year={2024}
}

@inproceedings{gao2017charades,
  title={Tall: Temporal activity localization via language query},
  author={Gao, Jiyang and Sun, Chen and Yang, Zhenheng and Nevatia, Ram},
  booktitle={Proceedings of the IEEE international conference on computer vision},
  pages={5267--5275},
  year={2017}
}

@inproceedings{rohrbach2014tacos,
  title={Coherent multi-sentence video description with variable level of detail},
  author={Rohrbach, Anna and Rohrbach, Marcus and Qiu, Wei and Friedrich, Annemarie and Pinkal, Manfred and Schiele, Bernt},
  booktitle={German conference on pattern recognition},
  pages={184--195},
  year={2014},
  organization={Springer}
}

@article{hu2025videommmu,
  title={Video-mmmu: Evaluating knowledge acquisition from multi-discipline professional videos},
  author={Hu, Kairui and Wu, Penghao and Pu, Fanyi and Xiao, Wang and Zhang, Yuanhan and Yue, Xiang and Li, Bo and Liu, Ziwei},
  journal={arXiv preprint arXiv:2501.13826},
  year={2025}
}

@inproceedings{zhao2025mmvu,
  title={Mmvu: Measuring expert-level multi-discipline video understanding},
  author={Zhao, Yilun and Zhang, Haowei and Xie, Lujing and Hu, Tongyan and Gan, Guo and Long, Yitao and Hu, Zhiyuan and Chen, Weiyuan and Li, Chuhan and Xu, Zhijian and others},
  booktitle={Proceedings of the CVPR},
  pages={8475--8489},
  year={2025}
}

@article{zhu2025mmr,
  title={MMR-V: What's Left Unsaid? A Benchmark for Multimodal Deep Reasoning in Videos},
  author={Zhu, Kejian and Jin, Zhuoran and Yuan, Hongbang and Li, Jiachun and Tu, Shangqing and Cao, Pengfei and Chen, Yubo and Liu, Kang and Zhao, Jun},
  journal={arXiv preprint arXiv:2506.04141},
  year={2025}
}

@inproceedings{han2025videoespresso,
  title={Videoespresso: A large-scale chain-of-thought dataset for fine-grained video reasoning via core frame selection},
  author={Han, Songhao and Huang, Wei and Shi, Hairong and Zhuo, Le and Su, Xiu and Zhang, Shifeng and Zhou, Xu and Qi, Xiaojuan and Liao, Yue and Liu, Si},
  booktitle={Proceedings of the CVPR},
  pages={26181--26191},
  year={2025}
}

@inproceedings{fu2025videomme,
  title={Video-mme: The first-ever comprehensive evaluation benchmark of multi-modal llms in video analysis},
  author={Fu, Chaoyou and Dai, Yuhan and Luo, Yongdong and Li, Lei and Ren, Shuhuai and Zhang, Renrui and Wang, Zihan and Zhou, Chenyu and Shen, Yunhang and Zhang, Mengdan and others},
  booktitle={Proceedings of the CVPR},
  pages={24108--24118},
  year={2025}
}

@article{wang2024lvbench,
  title={Lvbench: An extreme long video understanding benchmark},
  author={Wang, Weihan and He, Zehai and Hong, Wenyi and Cheng, Yean and Zhang, Xiaohan and Qi, Ji and Gu, Xiaotao and Huang, Shiyu and Xu, Bin and Dong, Yuxiao and others},
  journal={arXiv preprint arXiv:2406.08035},
  year={2024}
}

@article{wu2024longvideobench,
  title={Longvideobench: A benchmark for long-context interleaved video-language understanding},
  author={Wu, Haoning and Li, Dongxu and Chen, Bei and Li, Junnan},
  journal={NeurIPS},
  volume={37},
  pages={28828--28857},
  year={2024}
}

@article{zhou2024mlvu,
  title={Mlvu: A comprehensive benchmark for multi-task long video understanding},
  author={Zhou, Junjie and Shu, Yan and Zhao, Bo and Wu, Boya and Xiao, Shitao and Yang, Xi and Xiong, Yongping and Zhang, Bo and Huang, Tiejun and Liu, Zheng},
  journal={arXiv e-prints},
  pages={arXiv--2406},
  year={2024}
}

@article{ma2025videoeval,
  title={Videoeval-pro: Robust and realistic long video understanding evaluation},
  author={Ma, Wentao and Ren, Weiming and Jia, Yiming and Li, Zhuofeng and Nie, Ping and Zhang, Ge and Chen, Wenhu},
  journal={arXiv preprint arXiv:2505.14640},
  year={2025}
}

@inproceedings{li2022musicavqa,
  title={Learning to answer questions in dynamic audio-visual scenarios},
  author={Li, Guangyao and Wei, Yake and Tian, Yapeng and Xu, Chenliang and Wen, Ji-Rong and Hu, Di},
  booktitle={Proceedings of the CVPR},
  pages={19108--19118},
  year={2022}
}

@inproceedings{geng2025longvale,
  title={Longvale: Vision-audio-language-event benchmark towards time-aware omni-modal perception of long videos},
  author={Geng, Tiantian and Zhang, Jinrui and Wang, Qingni and Wang, Teng and Duan, Jinming and Zheng, Feng},
  booktitle={Proceedings of the CVPR},
  pages={18959--18969},
  year={2025}
}

@article{hong2025worldsense,
  title={Worldsense: Evaluating real-world omnimodal understanding for multimodal llms},
  author={Hong, Jack and Yan, Shilin and Cai, Jiayin and Jiang, Xiaolong and Hu, Yao and Xie, Weidi},
  journal={arXiv preprint arXiv:2502.04326},
  year={2025}
}

@article{mangalam2023egoschema,
  title={Egoschema: A diagnostic benchmark for very long-form video language understanding},
  author={Mangalam, Karttikeya and Akshulakov, Raiymbek and Malik, Jitendra},
  journal={NeurIPS},
  volume={36},
  pages={46212--46244},
  year={2023}
}

@article{liu2025shotbench,
  title={ShotBench: Expert-Level Cinematic Understanding in Vision-Language Models},
  author={Liu, Hongbo and He, Jingwen and Jin, Yi and Zheng, Dian and Dong, Yuhao and Zhang, Fan and Huang, Ziqi and He, Yinan and Li, Yangguang and Chen, Weichao and others},
  journal={arXiv preprint arXiv:2506.21356},
  year={2025}
}

@article{lei2025godbench,
  title={GODBench: A Benchmark for Multimodal Large Language Models in Video Comment Art},
  author={Lei, Yiming and Zhang, Chenkai and Liu, Zeming and Leng, Haitao and Liu, Shaoguo and Gao, Tingting and Liu, Qingjie and Wang, Yunhong},
  journal={arXiv preprint arXiv:2505.11436},
  year={2025}
}

@article{team2025kwai,
  title={Kwai Keye-VL Technical Report},
  author={Team, Kwai Keye and Yang, Biao and Wen, Bin and Liu, Changyi and Chu, Chenglong and Song, Chengru and Rao, Chongling and Yi, Chuan and Li, Da and Zang, Dunju and others},
  journal={arXiv preprint arXiv:2507.01949},
  year={2025}
}

@article{jia2025omnivqa,
  title={Scaling-up Perceptual Video Quality Assessment},
  author={Jia, Ziheng and Zhang, Zicheng and Zhang, Zeyu and Liang, Yingji and Zhu, Xiaorong and Li, Chunyi and Han, Jinliang and Wu, Haoning and Wang, Bin and Zhang, Haoran and others},
  journal={arXiv preprint arXiv:2505.22543},
  year={2025}
}

@article{comanici2025gemini25,
  title={Gemini 2.5: Pushing the frontier with advanced reasoning, multimodality, long context, and next generation agentic capabilities},
  author={Comanici, Gheorghe and Bieber, Eric and Schaekermann, Mike and Pasupat, Ice and Sachdeva, Noveen and Dhillon, Inderjit and Blistein, Marcel and Ram, Ori and Zhang, Dan and Rosen, Evan and others},
  journal={arXiv preprint arXiv:2507.06261},
  year={2025}
}

@misc{anthropic3Claude,
  title={Introducing the next generation of Claude, 2024},
  author={Anthropic},
  note={\url{https://www.anthropic.com/news/claude-3-5-sonnet}},
  year={2024}
}

@article{guo2025seed15vl,
  title={Seed1. 5-vl technical report},
  author={Guo, Dong and Wu, Faming and Zhu, Feida and Leng, Fuxing and Shi, Guang and Chen, Haobin and Fan, Haoqi and Wang, Jian and Jiang, Jianyu and Wang, Jiawei and others},
  journal={arXiv preprint arXiv:2505.07062},
  year={2025}
}

@article{zhang2023videollama,
  title={Video-llama: An instruction-tuned audio-visual language model for video understanding},
  author={Zhang, Hang and Li, Xin and Bing, Lidong},
  journal={arXiv preprint arXiv:2306.02858},
  year={2023}
}

@inproceedings{ren2024timechat,
  title={Timechat: A time-sensitive multimodal large language model for long video understanding},
  author={Ren, Shuhuai and Yao, Linli and Li, Shicheng and Sun, Xu and Hou, Lu},
  booktitle={Proceedings of the CVPR},
  pages={14313--14323},
  year={2024}
}

@inproceedings{wang2024internvideo2,
  title={Internvideo2: Scaling foundation models for multimodal video understanding},
  author={Wang, Yi and Li, Kunchang and Li, Xinhao and Yu, Jiashuo and He, Yinan and Chen, Guo and Pei, Baoqi and Zheng, Rongkun and Wang, Zun and Shi, Yansong and others},
  booktitle={European Conference on Computer Vision},
  pages={396--416},
  year={2024},
  organization={Springer}
}

@article{maaz2023video-chatgpt,
  title={Video-chatgpt: Towards detailed video understanding via large vision and language models},
  author={Maaz, Muhammad and Rasheed, Hanoona and Khan, Salman and Khan, Fahad Shahbaz},
  journal={arXiv preprint arXiv:2306.05424},
  year={2023}
}

@misc{zhang2024llavanextvideo,
  title={LLaVA-NeXT: A Strong Zero-shot Video Understanding Model},
  url={https://llava-vl.github.io/blog/2024-04-30-llava-next-video/},
  author={Zhang, Yuanhan and Li, Bo and Liu, haotian and Lee, Yong jae and Gui, Liangke and Fu, Di and Feng, Jiashi and Liu, Ziwei and Li, Chunyuan},
  month={April},
  year={2024}
}

@inproceedings{liu2025nvila,
  title={Nvila: Efficient frontier visual language models},
  author={Liu, Zhijian and Zhu, Ligeng and Shi, Baifeng and Zhang, Zhuoyang and Lou, Yuming and Yang, Shang and Xi, Haocheng and Cao, Shiyi and Gu, Yuxian and Li, Dacheng and others},
  booktitle={Proceedings of the CVPR},
  pages={4122--4134},
  year={2025}
}

@inproceedings{
zeng2025timesuite,
title={TimeSuite: Improving {MLLM}s for Long Video Understanding via Grounded Tuning},
author={Xiangyu Zeng and Kunchang Li and Chenting Wang and Xinhao Li and Tianxiang Jiang and Ziang Yan and Songze Li and Yansong Shi and Zhengrong Yue and Yi Wang and Yali Wang and Yu Qiao and Limin Wang},
booktitle={The Thirteenth International Conference on Learning Representations},
year={2025},
url={https://openreview.net/forum?id=nAVejJURqZ}
}

@article{zhang2025videollama3,
  title={Videollama 3: Frontier multimodal foundation models for image and video understanding},
  author={Zhang, Boqiang and Li, Kehan and Cheng, Zesen and Hu, Zhiqiang and Yuan, Yuqian and Chen, Guanzheng and Leng, Sicong and Jiang, Yuming and Zhang, Hang and Li, Xin and others},
  journal={arXiv preprint arXiv:2501.13106
        
        },
  year={2025}
}

@article{zhang2024llavavideo,
  title={Video instruction tuning with synthetic data},
  author={Zhang, Yuanhan and Wu, Jinming and Li, Wei and Li, Bo and Ma, Zejun and Liu, Ziwei and Li, Chunyuan},
  journal={arXiv preprint arXiv:2410.02713
        
        },
  year={2024}
}

@article{yao2024minicpm,
  title={MiniCPM-V: A GPT-4V Level MLLM on Your Phone},
  author={Yao, Yuan and Yu, Tianyu and Zhang, Ao and Wang, Chongyi and Cui, Junbo and Zhu, Hongji and Cai, Tianchi and Li, Haoyu and Zhao, Weilin and He, Zhihui and others},
  journal={arXiv preprint arXiv:2408.01800
        
        },
  year={2024}
}

@article{chen2024internvl25,
  title={Expanding performance boundaries of open-source multimodal models with model, data, and test-time scaling},
  author={Chen, Zhe and Wang, Weiyun and Cao, Yue and Liu, Yangzhou and Gao, Zhangwei and Cui, Erfei and Zhu, Jinguo and Ye, Shenglong and Tian, Hao and Liu, Zhaoyang and others},
  journal={arXiv preprint arXiv:2412.05271
        
        },
  year={2024}
}

@article{cheng2024videollama2,
  title={Videollama 2: Advancing spatial-temporal modeling and audio understanding in video-llms},
  author={Cheng, Zesen and Leng, Sicong and Zhang, Hang and Xin, Yifei and Li, Xin and Chen, Guanzheng and Zhu, Yongxin and Zhang, Wenqi and Luo, Ziyang and Zhao, Deli and others},
  journal={arXiv preprint arXiv:2406.07476
        
        },
  year={2024}
}

@article{team2025keye,
  title={Kwai Keye-VL Technical Report},
  author={Team, Kwai Keye and Yang, Biao and Wen, Bin and Liu, Changyi and Chu, Chenglong and Song, Chengru and Rao, Chongling and Yi, Chuan and Li, Da and Zang, Dunju and others},
  journal={arXiv preprint arXiv:2507.01949
        
        
        
        },
  year={2025}
}

@article{liu2025ola,
  title={Ola: Pushing the frontiers of omni-modal language model with progressive modality alignment},
  author={Liu, Zuyan and Dong, Yuhao and Wang, Jiahui and Liu, Ziwei and Hu, Winston and Lu, Jiwen and Rao, Yongming},
  journal={arXiv e-prints},
  pages={arXiv--2502},
  year={2025}
}

@article{chen2024internvl2,
    title={How Far Are We to GPT-4V? Closing the Gap to Commercial Multimodal Models with Open-Source Suites},
    author={Chen, Zhe and Wang, Weiyun and Tian, Hao and Ye, Shenglong and Gao, Zhangwei and Cui, Erfei and Tong, Wenwen and Hu, Kongzhi and Luo, Jiapeng and Ma, Zheng and others},
    journal={arXiv preprint arXiv:2404.16821
        
        },
    year={2024}
  }

@article{xu2025qwen25omni,
  title={Qwen2. 5-omni technical report},
  author={Xu, Jin and Guo, Zhifang and He, Jinzheng and Hu, Hangrui and He, Ting and Bai, Shuai and Chen, Keqin and Wang, Jialin and Fan, Yang and Dang, Kai and others},
  journal={arXiv preprint arXiv:2503.20215
        
        },
  year={2025}
}

@article{rasheed2025videomathqa,
  title={VideoMathQA: Benchmarking Mathematical Reasoning via Multimodal Understanding in Videos},
  author={Rasheed, Hanoona and Shaker, Abdelrahman and Tang, Anqi and Maaz, Muhammad and Yang, Ming-Hsuan and Khan, Salman and Khan, Fahad Shahbaz},
  journal={arXiv preprint arXiv:2506.05349
        
        
        
        },
  year={2025}
}

@inproceedings{li2023blip,
  title={Blip-2: Bootstrapping language-image pre-training with frozen image encoders and large language models},
  author={Li, Junnan and Li, Dongxu and Savarese, Silvio and Hoi, Steven},
  booktitle={International conference on machine learning},
  pages={19730--19742},
  year={2023},
  organization={PMLR}
}

@misc{seed16,
  title={Introduction to Techniques Used in {Seed1.6}},
  author={ByteDance},
  note={\url{https://seed.bytedance.com/en/seed1_6}},
  year={2025}
}

@article{lei2021qvhighlight,
  title={Detecting moments and highlights in videos via natural language queries},
  author={Lei, Jie and Berg, Tamara L and Bansal, Mohit},
  journal={NeurIPS},
  volume={34},
  pages={11846--11858},
  year={2021}
}

@article{lu2025vlmaspolicy,
  title={VLM as Policy: Common-Law Content Moderation Framework for Short Video Platform},
  author={Lu, Xingyu and Zhang, Tianke and Meng, Chang and Wang, Xiaobei and Wang, Jinpeng and Zhang, YiFan and Tang, Shisong and Liu, Changyi and Ding, Haojie and Jiang, Kaiyu and others},
  journal={arXiv preprint arXiv:2504.14904
        
        },
  year={2025}
}

\appendix
\section{Benchmark Details}

\subsection{Prompt Usage}

\paragraph{Prompt for video filtering} After obtaining the livestream video, we filtered out the overly simple ones. We used the proprietary model Seed1.5-VL to score each candidate video (on a scale of 1 to 10), and then removed the videos with a score less than 3.
\begin{center}
\noindent 
\begin{tabular}{|p{\textwidth}|}
\hline
You are an expert video analyst. Analyze the given video content and rate its cognitive complexity for generating high-level questions.

Evaluate the video on the following 4 dimensions (each on a scale from 1 to 10):

1. Content Depth: Does the video cover nuanced topics, abstract concepts, or multi-layered information?

2. Scene Dynamics: Are there multiple interacting visual elements, complex spatial relationships, or subtle contextual cues?

3. Narrative Structure: Does the video use non-linear storytelling, implicit reasoning, or sophisticated argumentative structure?

4. Cognitive Load: Does full understanding require sustained attention, working memory, inference, or background knowledge?

Scoring Instructions:
- Assign a score (1-10) to each dimension.
- Provide only the **final complexity score**, calculated as the average of the four scores, rounded to one decimal place.
- Your output format must be:

Complexity Score: X.X\\

\hline
\end{tabular}
\end{center}

\paragraph{Prompt for multi-agent system}
Previous work~\cite{han2025videoespresso} uses a single Multimodal Large Language Model (MLLM) to describe candidate videos. But that way introduces a large bias of a specific MLLM. To avoid this problem, we design a multi-agent system to comprehensively describe candidate videos. We designed each MLLM into four different roles based on its distinct characteristics. We design InternVL as a video scene and analysis expert, LLaVA-Video as a detailed analysis expert, Qwen2.5-VL as a logic and event analysis expert, and Seed1.5-VL as a knowledge expert.
\begin{center}
\noindent 
\begin{tabular}{|p{\textwidth}|}
\hline
\textbf{Instruction for InternVL3-38B: Video scene analysis expert} \\
You are a visual expert skilled at analyzing livestream video content. Please provide a high-level description of the video from a stylistic and structural perspective. Focus on the following aspects:

1. **Overall Theme**: What is the main theme of the video? (e.g., dancing, singing, chatting, talent show, product display, etc.)

2. **Scene and Background**: Is the scene set indoors, outdoors, or using a virtual background? Are there decorations, props, or lighting elements in the background?

3. **Visual Style**: Describe the visual tone—fresh, cool, retro, cute, or anime-style? Are any filters, beautification effects, background blurs, or overlays used?

4. **Video Quality and Composition**: Is the video clear and stable? Is the camera angle fixed or dynamic? Is the layout visually balanced?

5. **Streamer Appearance Overview (Non-Facial)**: How many people are in the scene? Describe their clothing style, accessories (e.g., hats, jewelry), or posture without focusing on facial expressions.

Write a concise and coherent paragraph summarizing this visual information.
\\ 
\textbf{Instruction for LLaVA-Video-72B: Video detail analysis expert} \\
You are an analyst focused on observing fine-grained details in livestream videos. Please describe the video content with attention to the following:

1. **Appearance Attributes (Non-Facial)**: Describe the streamer's outfit style, color scheme, body posture, accessories (e.g., hats, gloves, jewelry), or props they use.

2. **Talent or Activity**: Is the streamer performing a specific skill or activity (e.g., dancing, singing, drawing, cooking)? Is it a complex or trendy routine?

3. **Interaction Behavior**: Does the streamer respond to comments, acknowledge virtual gifts, or make gestures aimed at viewers?

4. **Multi-person Interaction**: Are multiple people present? Describe their roles or interactions (e.g., coordinated dancing, object handovers, spatial relation).

5. **Detail Cues**: Mention any noticeable items, gestures, or movements—like synchronized steps, use of props, or responses to virtual elements.

Avoid any mention of facial features or expressions. Focus on body-level actions and interactive clues.
\\
\hline
\end{tabular}

\end{center}

\clearpage

\begin{center}
    \noindent 
\begin{tabular}{|p{\textwidth}|}
\hline
\textbf{Instruction for Qwen2.5-VL-32B: Expert in logic and event analysis} \\
You are a specialist in temporal and causal reasoning for livestream videos. Analyze the logical structure and event flow of the video with a focus on:

1. **Event Sequence**: Identify the main events (e.g., starting a dance, receiving a gift, switching scenes) and their temporal order.

2. **Causal Relations**: Are there clear cause-effect relationships (e.g., gift received → change in movement or pose)? Explain the reasoning.

3. **Behavioral Change**: Is there any shift in physical action, posture, or camera usage? What seems to trigger those changes?

4. **Spatial Reasoning**: Do the participants change position? Are there any meaningful camera movements or scene transitions?

5. **Intentional Cues**: Does the streamer attempt to guide viewer behavior (e.g., asking for engagement, showing appreciation)? How is this intention expressed non-verbally?

Do not describe facial expressions or emotional states unless they are clearly reflected through full-body gestures or actions.
\\
\textbf{Instruction for Seed1.5-VL: Knowledge expert}\\
You are a cultural knowledge expert analyzing livestream video content for references to trends, media, and internet culture. Please address the following:

1. **Dance Recognition**: Does the performer execute a dance related to a viral trend or challenge? If identifiable, name the source or platform (e.g., TikTok).

2. **Media or Pop Culture References**: Are there references to known shows, anime, games, or internet memes via music, props, gestures, or outfits?

3. **Visual/Cultural Symbols**: Are there recognizable cultural elements (e.g., cosplay, virtual streamer style, fan service, idol culture)? Describe their appearance.

4. **Stickers and Effects**: Are visual effects or gift animations used (e.g., rockets, flying cars, emojis)? Identify and describe them.

5. **Music and Lyrics**: Is there music playing or lyrics displayed? If the song is known, please identify it and its cultural relevance.

You should avoid any assumption based on facial identity or expression. Focus instead on auditory, body-level, visual, and symbolic cultural cues.
\\
\hline
\end{tabular}
\end{center}

\paragraph{Prompt for asking questions} We use the following prompt to encourage a proprietary model, Doubao-Seed-1.6, to raise candidate questions for each task, inspired by our predefined seed question library.
\begin{center}
    \noindent 
\begin{tabular}{|p{\textwidth}|}
\hline
\textbf{Instruction for asking questions}\\
The above is a long live video. You need to ask questions about the ``\{Task Name\}" aspect of the long video. Here are some seed questions you can refer to:

\{Seed question library\}

If there are some more appropriate other video timing clues, you can also ask some other questions related to "\{Task Name\}".

You need to ensure that the question is related to the video, reasonable, and has a certain degree of difficulty and challenge. The question must be answered through multi-step reasoning.

Please generate a Chinese question with multiple options and only one correct answer, and provide a concise standard answer. Output in the following format, with only two lines of content:

question: xxx

answer: xxx
\\
\hline
\end{tabular}
\end{center}

\paragraph{Prompt for filter training videos} We use the following prompt to encourage a proprietary model to judge the distinct scenes and filter videos that only contain 1 scene.
\begin{center}
    \noindent 
\begin{tabular}{|p{\textwidth}|}
\hline
\textbf{Instruction to determine the number of scenes}\\
Given this video, analyze how many distinct scenes it contains. 
A scene is a segment with consistent visual setting.
Only return the number of scenes as an integer. Do not include any explanation.
\\
\hline
\end{tabular}
\end{center}

\paragraph{Prompt for generating training data} We use the following prompt to encourage a proprietary model to generate the synthetic training data automatically.
\begin{center}
    \noindent 
\begin{tabular}{|p{\textwidth}|}
\hline
\textbf{Instruction for generate training data}\\
Input a live video and you need to ask a question about the video content and provide an answer.
The question you raised needs to focus on spatial reasoning in the video. Perceive the spatial relationships among the instances observed in the video scene.

\{Seed question library\}

In addition to the given reference questions, you can also raise other related and more difficult questions.
Output the questions and answers in two lines:

question:

answer:
\\
\hline
\end{tabular}
\end{center}

\paragraph{Evaluation prompt} We use the following template to evaluate all models.
\begin{center}
    \noindent 
\begin{tabular}{|p{\textwidth}|}
\hline
\textbf{Evaluation template}\\
Comment: \\
ASR: \\
\{Question\}\\
\{Options\}\\

Please only output the letters of the correct options
\\
\hline
\end{tabular}
\end{center}

\subsection{Seed Question}
\paragraph{Seed question example} We list used some examples of seed question library in Table.~\ref{tab:seedquestion}.
\begin{table*}[h]
    \centering
    \begin{tabular}{l|p{0.8\textwidth}}
    \toprule
    \textbf{Task} & \textbf{Example} \\
    \midrule
     \emph{Video Topic} & What main emotions or moods (such as happiness or nostalgia) are conveyed by the dance movements of the people in the video and the overall atmosphere of the live broadcast?  \\
     \emph{Scene Recognition}  & List the background props that appear in the video \\
     \emph{Visual Style} & Which visual features best define the style of live streaming, such as dynamic camera movement and static composition? \\
     \emph{Quality Assessment} & Which split-screen pane stands out the most in terms of dynamic visual elements during live streaming? \\
     \midrule
     \emph{Detail Recognition} & What specific text is prominently displayed on the background sign or banner during a live stream? \\

     \emph{Attribute Recognition} & In the split-screen section, which clothing details stand out the most, such as patterns, colors or unique accessories? \\

     \emph{Event Recognition} & What is the specific prop that the main performer uses most frequently in a live stream, and what actions do they mainly perform with it? \\

     \emph{Talent Recognition} & How can the dancers' movements be coordinated with the rhythm of the background music to enhance the performance effect? \\

     \emph{Temporal Querying} & How does the background lighting change over time? \\

     \emph{Multi-person Interaction} & How many people are actively interacting in the scene\\

     \midrule

     \emph{Object Reasoning} & What is the nature of the relationship or interaction between [specific individuals/elements] in the video? \\
     \emph{Temporal Reasoning} & What is the correct sequence of [events, steps, tools, operations, products, etc.] presented in the video? \\
     \emph{Spatial Reasoning} & Which locations, regions or scenes are depicted or mentioned in the video? \\
     \emph{Causal Reasoning} & Do the depicted items/clothing match the specific situation (such as activities, events)? What's the reason? \\
     
    \bottomrule
    \end{tabular}
    \caption{Seed Question Example}
    \label{tab:seedquestion}
\end{table*}

\subsection{Word clouds}
\begin{figure*}
    \centering
    \includegraphics[width=0.8\linewidth]{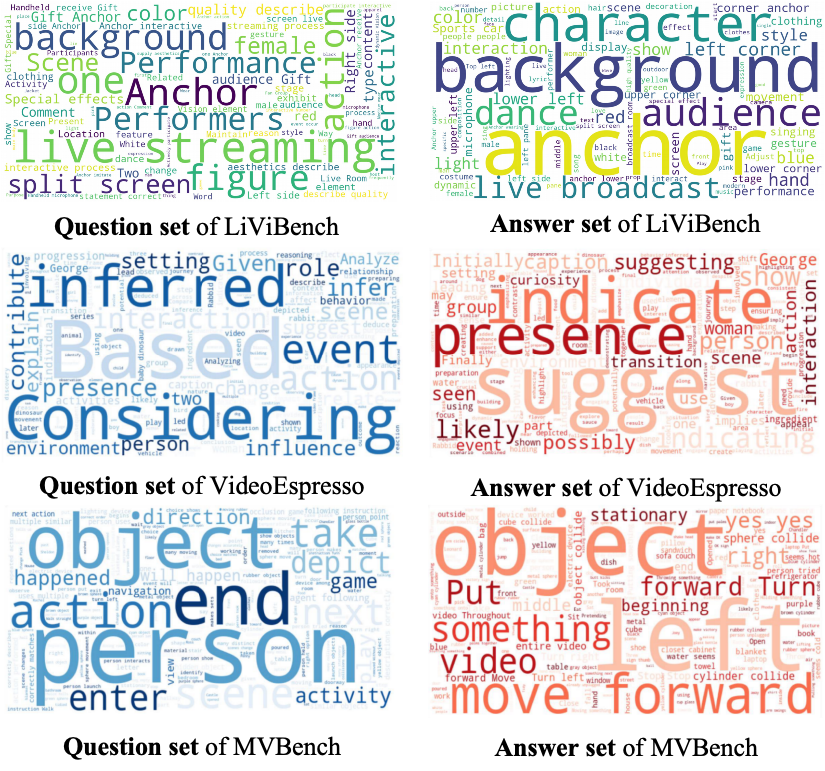}
    \caption{Comparison of word clouds for different benchmarks.}
    \label{fig:append_word_cloud}
\end{figure*}

We compare the word clouds with two general video benchmarks, including VideoEspresso~\cite{han2025videoespresso} and MVBench~\cite{li2024mvbench} in Figure.~\ref{fig:append_word_cloud}. The general benchmarks focus on more general items such as ``event", ``person", ``object". While our benchmark focuses on the livestream domain and interactive features, such as ``anchor". ``audience", ``performance", ``interactive". This comparison reveals 1) the uniqueness of our benchmark that distinguishes it from the general benchmark; 2) highlights the interactive features in the livestream domain.

\subsection{Benchmark preview}

\begin{figure*}
    \centering
    \includegraphics[width=\linewidth]{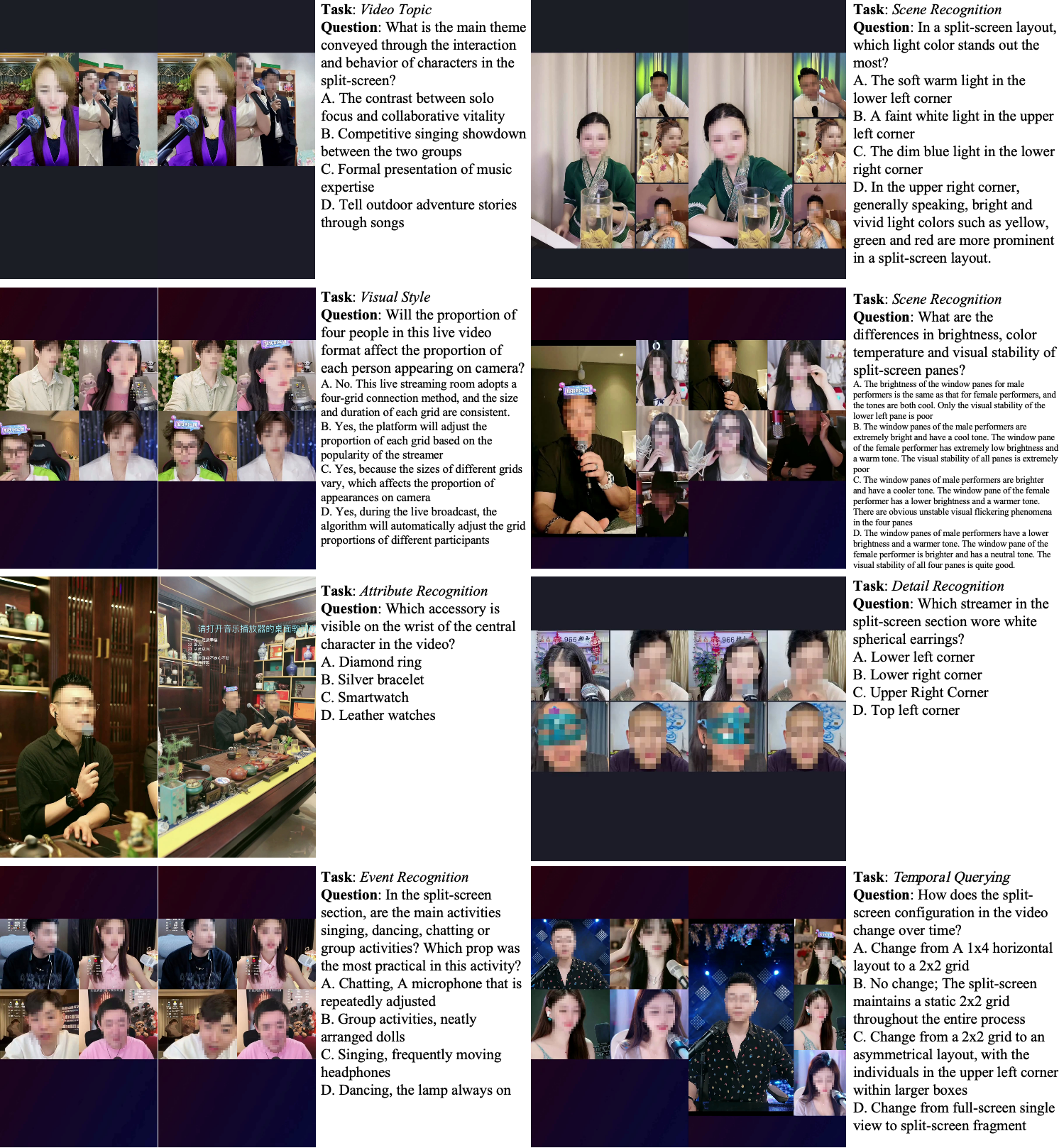}
    \caption{Benchmark preview.}
    \label{fig:preview1}
\end{figure*}

\begin{figure*}
    \centering
    \includegraphics[width=\linewidth]{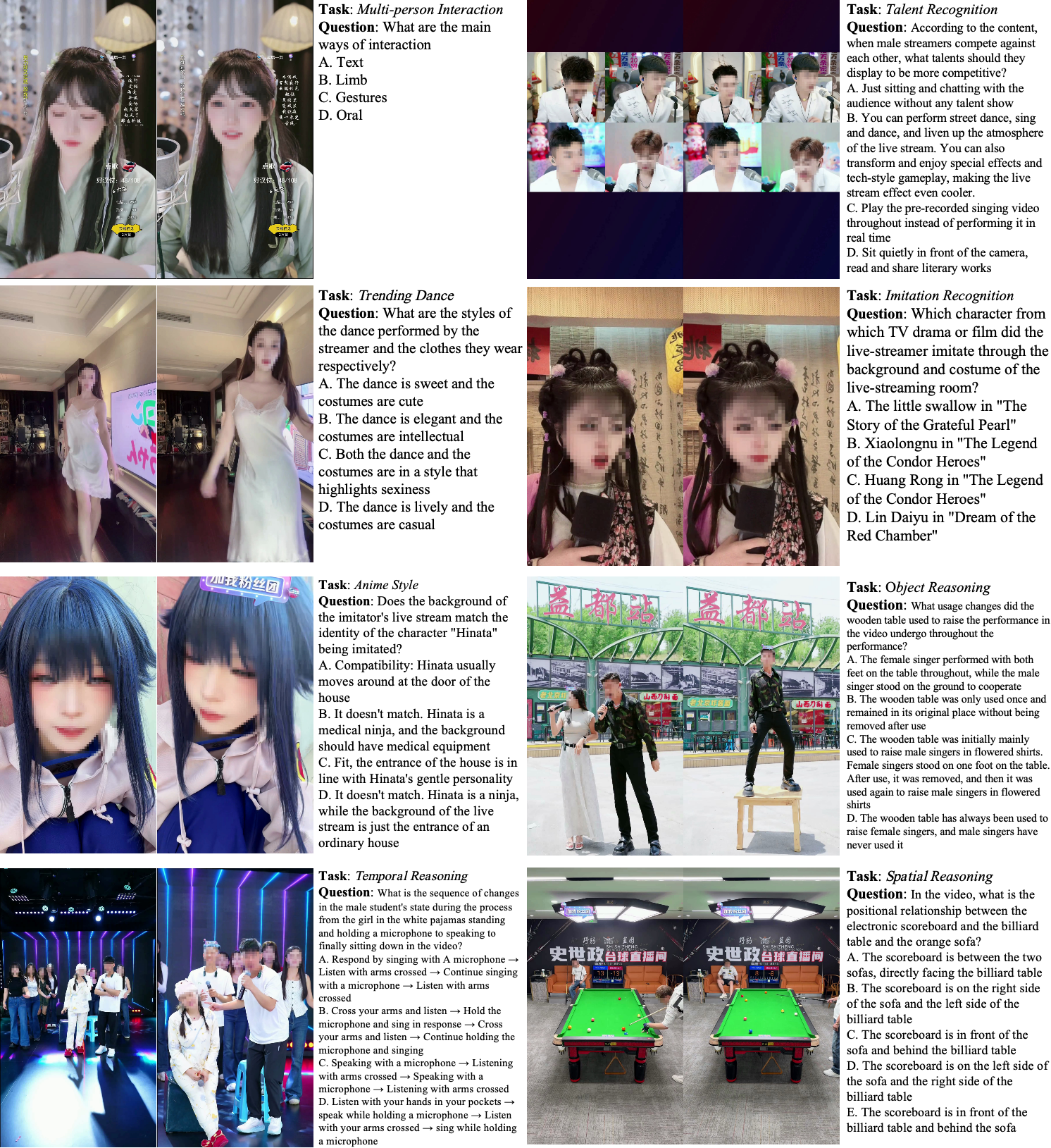}
    \caption{Benchmark preview.}
    \label{fig:preview2}
\end{figure*}

\begin{figure*}
    \centering
    \includegraphics[width=\linewidth]{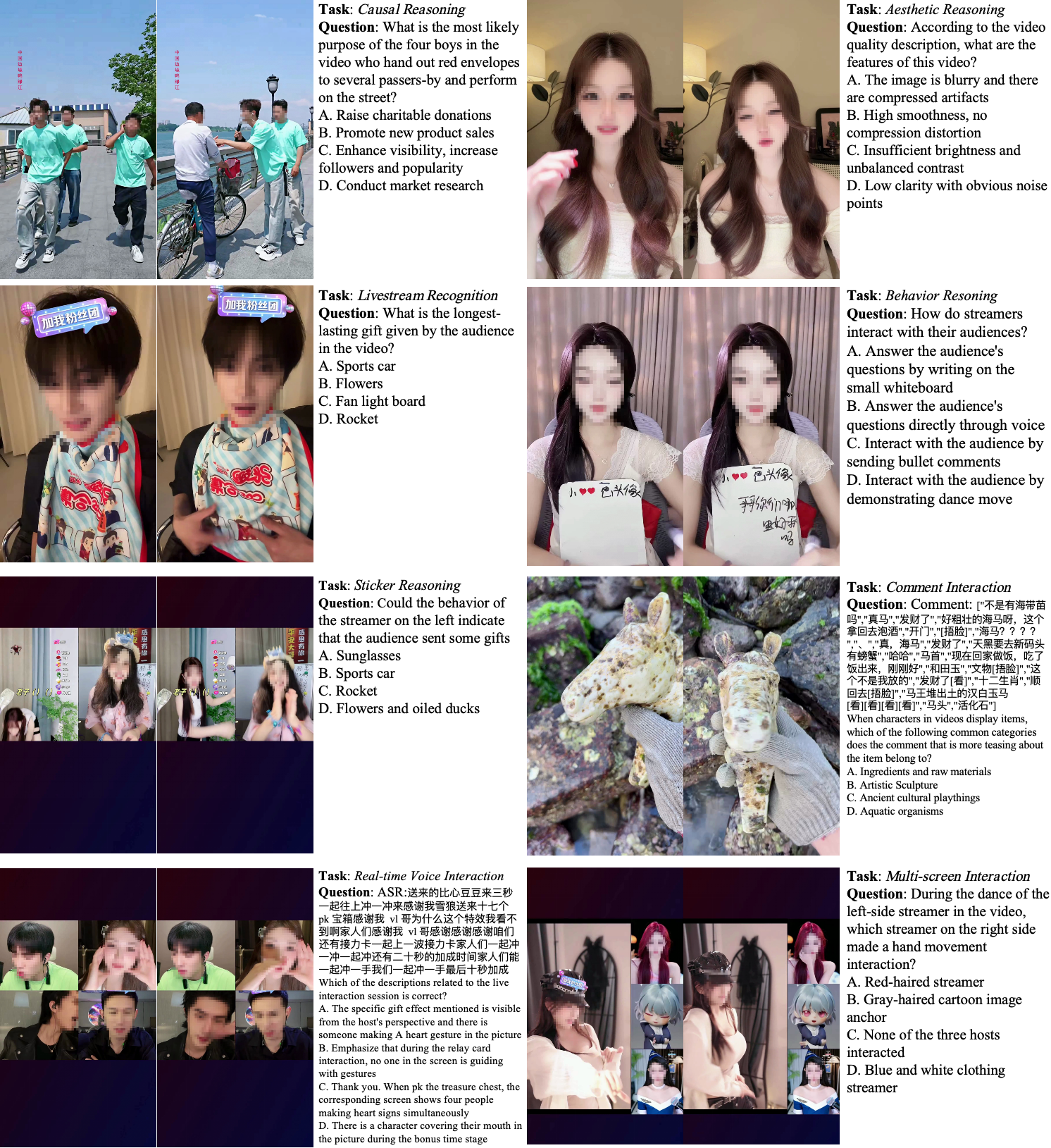}
    \caption{Benchmark preview.}
    \label{fig:preview3}
\end{figure*}

We show some examples for all 24 tasks in our benchmark in Figure.~\ref{fig:preview1}, Figure.~\ref{fig:preview2}, and Figure.~\ref{fig:preview3}. It should be noted that when evaluating each model, the input questions and options are all in Chinese, and the text of the comments and ASR is also in Chinese. All the models involved in the evaluation support Chinese input. All personal information (including private information such as facial features) will be desensitized during the assessment.

\end{document}